\documentclass[journal]{IEEEtran}

\usepackage{amsmath,amsfonts}
\usepackage{algpseudocode}
\usepackage{algorithmicx,algorithm}
\usepackage{hyperref}
\usepackage{array}
\usepackage[caption=false,font=normalsize,labelfont=sf,textfont=sf]{subfig}

\usepackage{textcomp}
\usepackage{stfloats}
\usepackage{url}
\usepackage{verbatim}
\usepackage{graphicx}
\usepackage{cite}

\usepackage[square,sort&compress,numbers]{natbib}
\usepackage{amsmath}
\usepackage{amsfonts} 
\usepackage{amssymb} 
\usepackage{multirow}
\usepackage{threeparttable}
\usepackage{booktabs}

\usepackage{bbm, dsfont}
\usepackage{bbding}
\usepackage{ifsym}

\graphicspath{{figs/}}
\graphicspath{{bio/}}

\begin{document}
\title{RSPrompter: Learning to Prompt for Remote Sensing Instance Segmentation based on Visual Foundation Model}

\author{
Keyan~Chen$^1$,~Chenyang~Liu$^1$,~Hao~Chen$^2$,~Haotian~Zhang$^1$,~Wenyuan~Li$^3$, Zhengxia~Zou$^1$,~and~Zhenwei~Shi$^{1, \star}$ \\
\vspace{4pt}
$^1$Beihang University, $^2$Shanghai AI Laboratory, $^3$The University of Hong Kong
}

\maketitle

\begin{abstract}

Leveraging the extensive training data from SA-1B, the Segment Anything Model (SAM) demonstrates remarkable generalization and zero-shot capabilities. However, as a category-agnostic instance segmentation method, SAM heavily relies on prior manual guidance, including points, boxes, and coarse-grained masks. Furthermore, its performance in remote sensing image segmentation tasks remains largely unexplored and unproven.
In this paper, we aim to develop an automated instance segmentation approach for remote sensing images, based on the foundational SAM model and incorporating semantic category information. Drawing inspiration from prompt learning, we propose a method to learn the generation of appropriate prompts for SAM. This enables SAM to produce semantically discernible segmentation results for remote sensing images, a concept we have termed RSPrompter. We also propose several ongoing derivatives for instance segmentation tasks, drawing on recent advancements within the SAM community, and compare their performance with RSPrompter.
Extensive experimental results, derived from the WHU building, NWPU VHR-10, and SSDD datasets, validate the effectiveness of our proposed method. The code for our method is publicly available at \url{https://kyanchen.github.io/RSPrompter}.

\end{abstract}


\IEEEpeerreviewmaketitle
\section{Introduction}

\IEEEPARstart{I}nstance segmentation is a pivotal task in the analysis of remote sensing images, facilitating a semantic-level understanding of each instance within the images. This process yields crucial information regarding the location (where), category (what), and shape (how) of each object \cite{su2020hq, su2019object, zhang2021semantic, xu2021improved, liu2021catnet, zou2023object, chen2021building, chen2022resolution, chen2022degraded}. The accurate perception and comprehension of surfaces in remote sensing images significantly propel the advancement of remote sensing for earth observation. The applications of this technology span a multitude of domains, including but not limited to national defense, land surveying, disaster monitoring, and traffic planning \cite{cheng2016survey, chen2021building, li2022geographical, li2020object, chen2023remote, chen2023continuous, chen2022contrastive}.

Deep learning algorithms have shown significant promise in instance segmentation for remote sensing images, demonstrating their capacity to extract deep, discernible features from raw data \cite{liu2018path, roscher2020semcity, minaee2021image, fan2022efficient}. Currently, the predominant instance segmentation algorithms include two-stage R-CNN series algorithms (\textit{e.g.}, Mask R-CNN \cite{he2017mask}, Cascade Mask R-CNN \cite{cai2019cascade}, Mask Scoring R-CNN \cite{huang2019mask}, HTC \cite{chen2019hybrid}, and HQ-ISNet \cite{su2020hq}), as well as one-stage algorithms (\textit{e.g.}, YOLACT \cite{bolya2019yolact}, BlendMask \cite{chen2020blendmask}, EmbedMask \cite{ying2019embedmask}, Condinst \cite{tian2020conditional}, SOLO \cite{wang2020solo}, and Mask2Former \cite{cheng2022masked}).
Nevertheless, the intricacy of remote sensing image backgrounds and the diversity of scenes pose challenges to the generalization and adaptability of these algorithms. Consequently, the development of instance segmentation models capable of accommodating broad remote sensing scenarios is of crucial importance for the interpretation of remote sensing imagery.

\begin{figure}[t]
\centering
\resizebox{\linewidth}{!}{
\includegraphics[width=\linewidth]{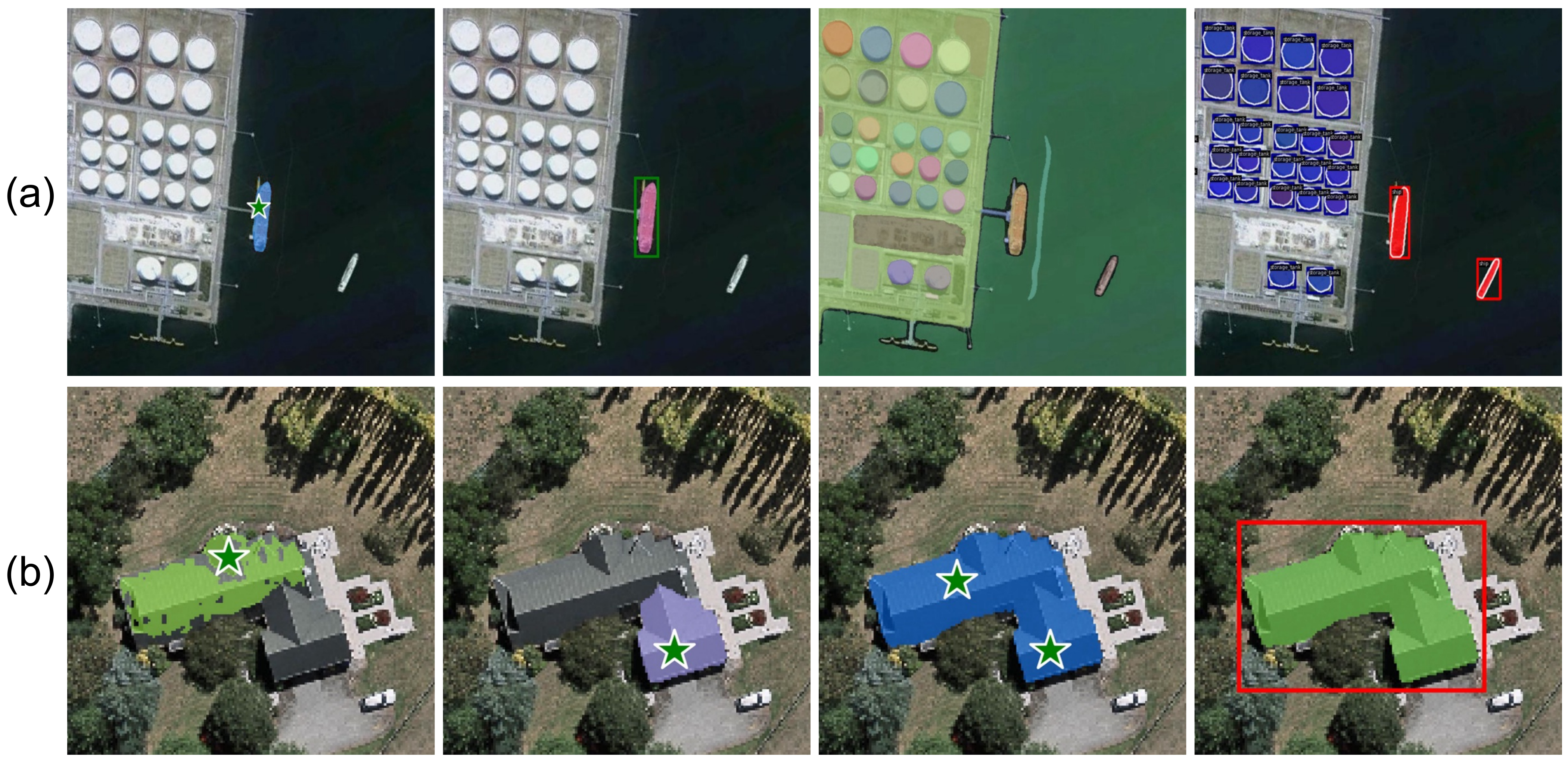}
}
\caption{
Comparative instance segmentation results. (a) demonstrates the instance segmentation results derived from point-based prompts, box-based prompts, SAM's ``everything" mode (which segments all objects within the image), and RSPrompter. Note that SAM executes category-agnostic instance segmentation, dependent on manually provided prior prompts. (b) showcases the segmentation results of point-based prompts originating from various locations, a two-point-based prompt, and a box-based prompt. It is evident that the type, location, and number of prompts significantly impact the results produced by SAM.}
\label{fig:teaser}
\end{figure}

In recent years, notable advancements have been made in foundational models, such as GPT-4 \cite{openai2023gpt4}, Flamingo \cite{alayrac2022flamingo}, and SAM \cite{kirillov2023segment}, which have significantly contributed to societal progress. Despite the inherent big data attributes of remote sensing since its inception \cite{li2021geographical, li2022geographical}, there has yet to be the development of foundational models specifically tailored to this field. The primary aim of this paper is not to create a universal foundational model for remote sensing but rather to investigate the applicability of the SAM foundational model, originating from the computer vision domain, to instance segmentation in remote sensing imagery. We foresee that such foundational models will catalyze continued advancement and growth within the remote sensing field.

The SAM model, trained on over a billion masks, demonstrates exceptional generalization capabilities, enabling it to segment any object in any image without additional training \cite{kirillov2023segment}. This advancement opens new horizons for intelligent image analysis and understanding \cite{ma2023segment, cen2023segment, zhang2023personalize, shaharabany2023autosam, osco2023segment, hu2023efficiently}. Nevertheless, SAM's interactive framework necessitates a prior prompt, such as a point, box, or mask, to be provided in conjunction with the input image. It operates as a category-agnostic segmentation method, as illustrated in Fig. \ref{fig:teaser} (a). These constraints render SAM unsuitable for the fully automatic understanding of remote-sensing images. Furthermore, we note that the complex background interference and the absence of well-defined object edges in remote sensing image scenarios pose significant challenges to SAM's segmentation capabilities. SAM's performance in segmenting remote-sensing image objects is heavily dependent on the type, location, and quantity of prompts. In most cases, refined manual prompts are indispensable to achieve the desired outcomes, as shown in Fig. \ref{fig:teaser} (b). This observation indicates that SAM suggests considerable limitations when directly applied to instance segmentation in remote-sensing images.

To enhance the remote sensing image instance segmentation of the foundation model, we propose a novel approach called RSPrompter, for learning how to generate prompts that can enhance the capabilities of the SAM framework.
Our research is primarily motivated by the SAM, wherein each group of prompts can yield an instance-specific mask via the mask decoder. We postulate that if we could autonomously generate groups of category-related prompts, the SAM decoder would be capable of generating multiple instance-level masks, each with their respective category labels. However, this process presents two significant challenges: (i) How can we source category-related prompts? (ii) What type of prompts should be selected for the mask decoder?

Our research is primarily centered on the SAM framework, a category-agnostic interactive segmentation model. We propose a lightweight feature enhancer to collect features from the SAM encoder's intermediate layers for the subsequent prompter. The prompter can generate prompts with semantic categories.
Furthermore, the prompts in the SAM model can take the form of points (foreground/background points), boxes, or masks. The generation of point coordinates necessitates a search within the original SAM prompt's manifold, which significantly constrains the prompter's optimization space. To circumvent this limitation, we propose a more flexible representation of prompts, \textit{i.e.}, prompt embeddings, rather than the original coordinates. Those can be interpreted as the embeddings of points or boxes.
This strategy also mitigates the challenge of gradient flow from high-dimensional to low-dimensional features and vice versa, \textit{i.e.}, from high-dimension image features to point coordinates and subsequently to positional encodings. 

In addition to our primary research (RSPrompter in Fig. \ref{fig:model_gallery} (d)), we have conducted an extensive review of the current advancements and derivatives in the SAM community \cite{zhang2023personalize, ke2023segment, ma2023segment}. These primarily include methods based on the SAM backbone, methods that integrate SAM with classifiers, and methods that combine SAM with detectors. The primary contributions of this paper can be summarized as follows:

\vspace{4pt}

(i) We propose a novel prompt learning method that augments the SAM model's capabilities, thereby facilitating instance segmentation in remote sensing imagery.

(ii) We undertake a comprehensive evaluation of the SAM model's performance when integrated with other discriminative models for instance segmentation pertaining to remote sensing imagery.

(iii) We present extensive results on three diverse remote sensing instance segmentation datasets, varying in size, categories, and modalities, to demonstrate the efficacy of the proposed RSPrompter.

The structure of this paper is organized as follows: Sec. II offers a thorough review of the relevant literature. Sec. III provides an in-depth exploration of the extension methods based on the SAM model for instance segmentation, as well as a detailed discussion of the proposed RSPrompter. Sec. IV presents both quantitative and qualitative results, supplemented by ablation studies to further substantiate our findings. Finally, Sec. V concludes the paper and encapsulates the key insights.

\section{Related Works}

\subsection{Deep Learning based Instance Segmentation}

The goal of instance segmentation is to pinpoint the location of each target within an image and provide a corresponding semantic mask. This task is inherently more complex than object detection and semantic segmentation \cite{minaee2021image, hafiz2020survey}. Current deep learning-based instance segmentation approaches can be broadly divided into two categories: two-stage and single-stage methods. The former primarily builds upon the Mask R-CNN \cite{he2017mask} series, which has evolved from the two-stage Faster R-CNN \cite{ren2015faster} object detector by incorporating a parallel mask prediction branch. As research progresses, a growing number of researchers are refining this framework to achieve enhanced performance. PANet \cite{liu2018path} optimizes the information path between features by introducing a bottom-up path based on FPN \cite{lin2017feature}. In HTC \cite{chen2019hybrid}, a multi-task, multi-stage hybrid cascade structure is proposed, and the spatial context is augmented by integrating the segmentation branch, resulting in significant performance improvements over Mask R-CNN and Cascade Mask R-CNN \cite{cai2019cascade}. The Mask Scoring R-CNN \cite{huang2019mask} incorporates a mask IoU branch within the Mask R-CNN framework to evaluate segmentation quality. The HQ-ISNet \cite{su2020hq} introduces an instance segmentation method for remote sensing imagery based on Cascade Mask R-CNN, which fully exploits multi-level feature maps and preserves the detailed information contained within high-resolution images.

While two-stage methods can produce refined segmentation results, achieving the desired speed of segmentation remains a challenge. With the increasing popularity of single-stage object detectors, numerous researchers have sought to adapt these detectors for instance segmentation tasks. For instance, YOLACT \cite{bolya2019yolact} addresses the instance segmentation task by generating a set of prototype masks and predicting mask coefficients for each instance. CondInst \cite{tian2020conditional} provides a novel perspective on the instance segmentation problem by utilizing a dynamic masking head, outperforming existing methods such as Mask R-CNN in terms of instance segmentation performance. SOLO \cite{wang2020solo} reframes the instance segmentation problem as predicting semantic categories and generating instance masks for each pixel in the feature map.
With the widespread use of Transformers \cite{vaswani2017attention}, DETR \cite{carion2020end} has emerged as a fully end-to-end object detector. Inspired by the task modeling and training procedures employed in DETR, MaskFormer \cite{cheng2021per} treats segmentation tasks as mask classification problems, but with a slow convergence speed. Mask2Former \cite{cheng2022masked} introduces masked attention to limit cross-attention to the foreground region, significantly enhancing network training speed.

Instance segmentation and object detection tasks are mutually reinforcing, and their development has reached a plateau at the technical level. Currently, research on foundational segmentation and detection models has become a popular area of focus \cite{kirillov2023segment, minderer2022simple, liang2023open, liu2023grounding, wang2023seggpt, zou2023segment}. In this paper, we validate the performance of the SAM foundation model when applied to instance segmentation tasks in remote sensing imagery.

\subsection{Foundation Model}

In recent years, foundation models have sparked a transformative shift in the field of artificial intelligence. Training on vast quantities of data has endowed these models with impressive zero-shot generalization capabilities across a multitude of scenarios \cite{radford2021learning, jia2021scaling, kirillov2023segment, sharif2014cnn}. Prominent models such as Chat-GPT \cite{ouyang2022training}, GPT-4 \cite{openai2023gpt4}, and Stable Diffusion \cite{rombach2022high} have further propelled the evolution of artificial intelligence, contributing significantly to the advancement of human civilization and exerting considerable influence across various industries. Motivated by the success of foundational models in Natural Language Processing (NLP), researchers have begun to investigate their potential applicability within the domain of computer vision. While the majority of these models aim to extract accessible knowledge from freely available data \cite{radford2021learning, alayrac2022flamingo, chen2023ovarnet}, the recent SAM model \cite{kirillov2023segment} takes an innovative approach by constructing a data engine in which the model is co-developed with model-in-the-loop dataset annotation. SAM uniquely utilizes an extensive collection of masks (11 million images, comprising over 1 billion masks), demonstrating robust generalization capabilities. However, it was originally designed as a task-agnostic segmentation model requiring prompts (\textit{i.e.}, input of prior points, bounding boxes, or masks), and thus, it does not facilitate end-to-end automated segmentation perception. In this paper, we do not focus on the design and training of a foundational remote sensing instance segmentation model. Instead, we investigate the potential of leveraging SAM's powerful general segmentation capabilities for remote-sensing images, with the aim of inspiring readers and fellow researchers. Moreover, the proposed learn-to-prompt method can be extended to other foundation visual models beyond SAM for task-specific or domain-specific downstream tasks.

\subsection{Prompt Learning}

Historically, machine learning tasks have predominantly centered on fully supervised learning, where task-specific models were trained exclusively on labeled instances of the target task \cite{krizhevsky2017imagenet, alom2018history}. Over time, the learning paradigm has significantly evolved, transitioning from fully supervised learning to a ``pre-training and fine-tuning" approach for downstream tasks. This shift allows models to leverage the general features acquired during pre-training \cite{russakovsky2015imagenet, deng2009imagenet, simonyan2014very, he2016deep}. In the recent era of foundation models, a new paradigm, ``pre-training and prompting" \cite{lester2021power, feng2022promptdet, chen2023ovarnet, jia2022visual, liu2023pre, zhou2022learning}, has emerged. In this paradigm, researchers no longer exclusively adapt the model to downstream tasks. Instead, they redesign the input using prompts to reconstruct downstream tasks to align with the original pre-training task \cite{radford2021learning, devlin2018bert, radford2019language}.
Prompt learning can aid in reducing semantic discrepancies, bridging the gap between pre-training and fine-tuning, and preventing overfitting of the head. Since the introduction of GPT-3 \cite{brown2020language}, prompt learning has evolved from traditional discrete \cite{liu2023pre} and continuous prompt construction \cite{zhou2022learning, chen2023ovarnet} to large-scale model-oriented in-context learning \cite{alayrac2022flamingo}, instruction-tuning \cite{liu2023visual, gupta2022improving, peng2023instruction}, and chain-of-thought approaches \cite{wei2022chain, wang2022self, zhang2022automatic}. Current methods for constructing prompts primarily involve manual templates, heuristic-based templates, generation, word embedding fine-tuning, and pseudo tokens \cite{liu2023pre, wang2022learning}. In this paper, we propose a prompt generator that generates SAM-compatible prompt inputs. This prompt generator is category-related and yields semantic instance segmentation results.
\section{Methodology}

In this section, we will introduce our proposed RSPrompter, a learning-to-prompt approach grounded in the SAM framework, specifically designed for instance segmentation in remote sensing imagery. The section will encompass the following aspects: a revisit of the SAM framework, the introduction of both anchor-based and query-based RSPrompter, including their loss functions, and some straightforward extensions adapting SAM to instance segmentation.

\begin{figure*}[t]
\centering
\resizebox{\linewidth}{!}{
\includegraphics[width=\linewidth]{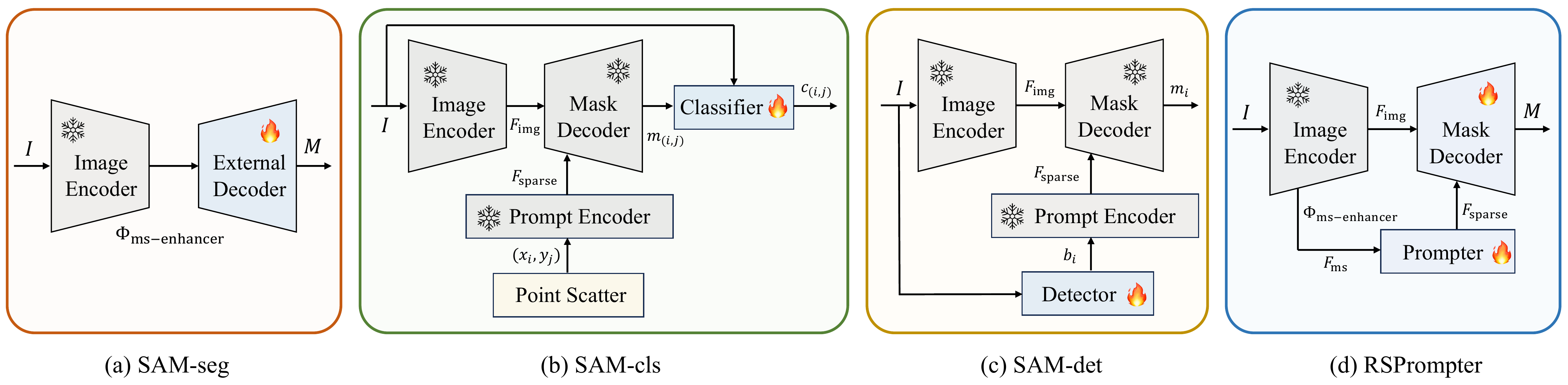}
}
\caption{From left to right, the figure illustrates SAM-seg, SAM-cls, SAM-det, and RSPrompter as alternative solutions for applying SAM to remote sensing image instance segmentation tasks.} (a) An instance segmentation head is added after SAM's image encoder. (b) SAM's ``everything" mode generates masks for all objects in an image, which are subsequently classified into specific categories by a classifier. (c) Object bounding boxes are first produced by an object detector and then used as prior prompts for SAM to obtain the corresponding masks. (d) The proposed RSPrompter creates category-relevant prompt embeddings for instant segmentation. The snowflake icon in the diagram indicates that the model parameters are maintained in a frozen state, while the fire symbol signifies active training.
\label{fig:model_gallery}
\end{figure*}

\begin{figure}[t]
\centering
\resizebox{0.99\linewidth}{!}{
\includegraphics[width=\linewidth]{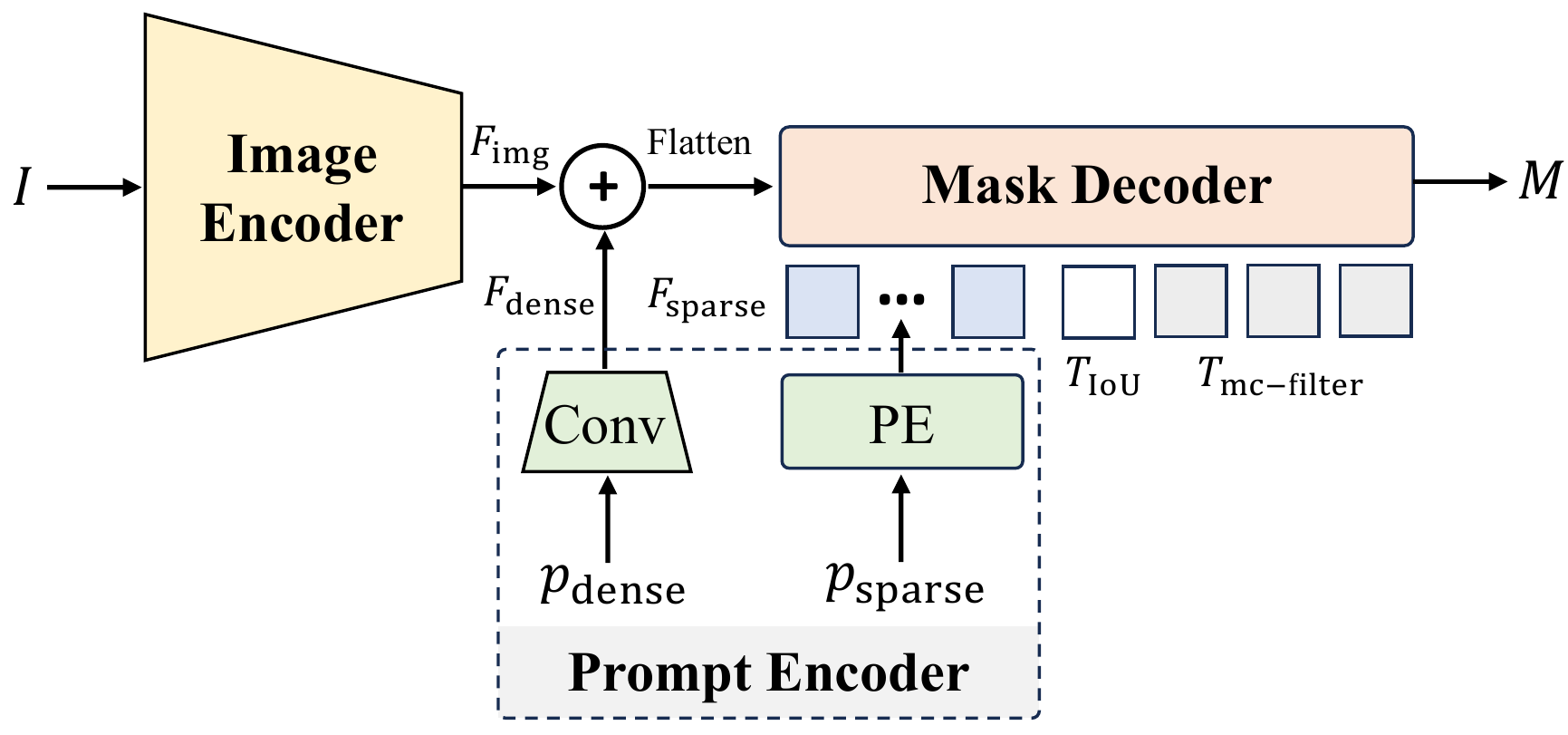}
}
\caption{The figure presents a schematic illustration of the SAM framework, encompassing an image encoder, a prompt encoder, and a mask decoder. The framework generates corresponding object masks with the given input prompts.}
\label{fig:sam}
\end{figure}

\subsection{A Revisit of SAM}

The SAM framework, an interactive segmentation approach predicated on provided prompts such as foreground/background points, bounding boxes, and masks, comprises three primary components: an image encoder ($\Phi_{\text{i-enc}}$), a prompt encoder ($\Phi_{\text{p-enc}}$), and a mask decoder ($\Phi_{\text{m-dec}}$). 
SAM employs a pre-trained Masked AutoEncoder (MAE) \cite{he2022masked} based on the Vision Transformer (ViT) \cite{dosovitskiy2020image} to process images into intermediate features, and encodes the prior prompts as embedding tokens. The mask decoder's cross-attention mechanism then enables interaction between image features and prompt embeddings, culminating in a mask output. This process is depicted in Fig. \ref{fig:sam} and can be expressed as follows:
\begin{align} 
\begin{split} 
F_{\text{img}} &= \Phi_{\text{i-enc}}(\mathcal{I}) \\
F_{\text{sparse}} &= \Phi_{\text{p-enc}}({p_{\text{sparse}}}) \\
F_{\text{dense}} &= \Phi_{\text{p-enc}}(p_{\text{dense}}) \\
F_{\text{out}} &= \text{Cat}(T_{\text{mc-filter}}, T_{\text{IoU}}, F_{\text{sparse}}) \\
\mathcal{M} &= \Phi_{\text{m-dec}}(F_{\text{img}} + F_{\text{dense}}, F_{\text{out}}) \\
\label{eq:overall} 
\end{split} 
\end{align}
where $\mathcal{I} \in \mathbb{R}^{H \times W \times 3}$ denotes the original image, $F_{\text{img}} \in \mathbb{R}^{h \times w \times c}$ represents the intermediate image features, ${p_{\text{sparse}}}$ encompasses the sparse prompts including foreground/background points and bounding boxes, and $F_{\text{sparse}} \in \mathbb{R}^{k \times c}$ signifies the sparse prompt tokens encoded by $\Phi_{\text{p-enc}}$. 
Furthermore, $p_{\text{dense}} \in \mathbb{R}^{H \times W}$ refers to the coarse segmentation mask, and $F_{\text{dense}} \in \mathbb{R}^{h \times w \times c}$ is the dense representation extracted by the prompter encoder $\Phi_{\text{p-enc}}$, which is an optional input for SAM. 
$T_{\text{mc-filter}} \in \mathbb{R}^{4 \times c}$ and $T_{\text{IoU}} \in \mathbb{R}^{1 \times c}$ are the pre-inserted learnable tokens representing four different mask filters and their corresponding IoU predictions. $\mathcal{M}$ denotes the predicted multi-choice masks. 
In our study, diverse outputs are not required, so we directly select the first mask as the final prediction.

\subsection{RSPrompter}
\subsubsection{Overview}

The structure of the proposed RSPrompter is depicted in Fig. \ref{fig:model_gallery} (d). Let us consider a training dataset, denoted as $\mathcal{D}_{\text{train}} = \{(\mathcal{I}_1, y_1), \cdots, (\mathcal{I}_N, y_N)\}$, wherein $\mathcal{I}_i \in \mathbb{R}^{H \times W \times 3}$ signifies an image, and $y_i = \{b_i, c_i, m_i\}$ corresponds to its respective ground-truth annotations, encompassing the coordinates of $n$ object bounding boxes ($b_i \in \mathbb{R}^{n_i \times 4}$), their affiliated semantic categories ($c_i \in \mathbb{R}^{n_i \times \mathcal{C}}$), and binary masks ($m_i \in \mathbb{R}^{n_i \times H \times W}$). The primary objective is to train a prompter for SAM that is capable of processing any image from a test set ($\mathcal{I}_k \sim \mathcal{D}_{\text{test}}$), concurrently localizing the objects and inferring their semantic categories and instance masks, which can be articulated as follows:
\begin{align} 
\begin{split} 
F_{\text{img}}, \{F_i\} &= \Phi_{\text{i-enc}}(\mathcal{I}_k) \\
\{F_{\text{ms}}^j\} &= \Phi_{\text{ms-enhancer}}(\{F_i\}) \\ 
\{F_{\text{sparse}}^m, c_m\} &= \Phi_{\text{prompter}}(\{F_{\text{ms}}^j\}) \\
F_{\text{out}} &= \text{Cat}(T_{\text{mc-filter}}, T_{\text{IoU}}, F_{\text{sparse}}^m) \\
\mathcal{M}_m &= \Phi_{\text{m-dec}}(F_{\text{img}}, F_{\text{out}}) \\ \label{eq:RSPrompter} 
\end{split} 
\end{align}
where the image ($\mathcal{I}_k \in \mathbb{R}^{H \times W \times 3}$) is processed by the frozen SAM image encoder to generate $F_{\text{img}} \in \mathbb{R}^{h \times w \times c}$ and multiple intermediate feature maps $\{F_i\} \in \mathbb{R}^{K \times h \times w \times c}$. $F_{\text{img}}$ is utilized by the SAM decoder to obtain prompt-guided masks, while $\{F_i\}$ is progressively processed by an efficient multi-scale feature enhancer ($\Phi_{\text{ms-enhancer}}$) to obtain the multi-scale feature maps ($\{F_{\text{ms}}^j \in \mathbb{R}^{\frac{H}{2^{j+1}} \times \frac{W}{2^{j+1}} \times c} \}$, $j \in \{1,2,3,4,5\}$) and a prompter ($\Phi_{\text{prompter}}$) to acquire multiple groups of prompts ($F_{\text{sparse}}^m \in \mathbb{R}^{K_p \times c}, m \in \{1, \cdots, N_p\}$) and corresponding semantic categories ($c_m \in \mathbb{R}^{\mathcal{C}}, m \in \{1, \cdots, N_p\}$). $K_p$ defines the number of prompt embeddings for each mask generation. $N_p$ is the number of prompt groups to define the number of output instance masks. Two distinct structures, namely anchor-based and query-based, have been employed for the prompt generator ($\Phi_{\text{prompter}}$).

It is crucial to note that $F_{\text{sparse}}^m$ only contains foreground target instance prompts, with the semantic category given by $c_m$. A single $F_{\text{sparse}}^m$ is a combination of multiple prompts, \textit{i.e.}, representing an instance mask with multiple point embeddings or a bounding box. For the sake of simplicity, the superscript $k$ in $\mathcal{I}_k$ will be omitted when describing the following proposed model.

\vspace{8pt}
\subsubsection{Multi-scale Feature Enhancer} \label{sec:ms-enhancer}

\begin{figure}[t]
\centering
\resizebox{\linewidth}{!}{
\includegraphics[width=\linewidth]{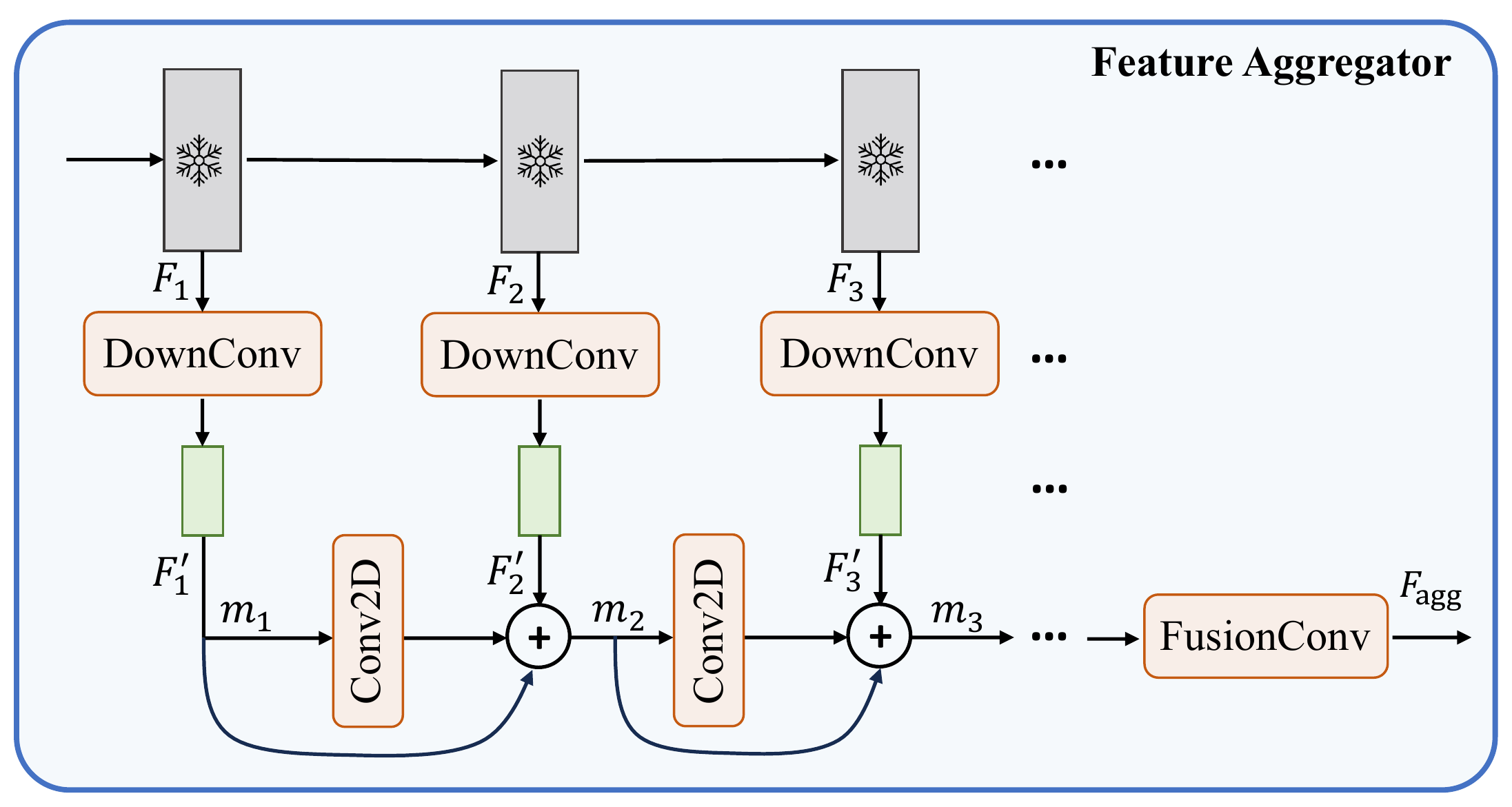}
}
\caption{
Illustration of the proposed lightweight feature aggregator, which extracts semantic information from the ViT backbone and performs a lightweight fusion process.
}
\label{fig:aggregator}
\end{figure}

To extract semantically pertinent and discriminative features without augmenting the computational complexity of the prompter, we propose a lightweight multi-scale feature enhancer. This enhancer comprises a feature aggregator and a feature splitter. The feature aggregator is designed to learn representative semantic features from a multitude of intermediate feature layers within the SAM ViT backbone. Concurrently, the feature splitter is employed to generate multi-scale pyramid feature maps.

\vspace{6pt}
\noindent \textbf{Feature Aggregator:} The feature aggregator, as illustrated in Fig. \ref{fig:aggregator}, can be defined recursively as follows:
\begin{align}
\begin{split}
    F_i^{\prime} &= \Phi_{\text{DownConv}}(F_i)
    \\
    m_1 &= F_1^{\prime}
    \\
    m_i &= m_{i-1} + \Phi_{\text{Conv2D}}(m_{i-1}) + F_i^{\prime}
    \\
    F_{\text{agg}} &= \Phi_{\text{FusionConv}}(m_k)
    \\
\end{split}
\end{align}
where $F_i \in \mathbb{R}^{h \times w \times c}$ and $F_i^{\prime} \in \mathbb{R}^{h \times w \times 32}$ represent the SAM backbone's features and down-sampled features produced by $\Phi_{\text{DownConv}}$, respectively. The process involves the utilization of a $1 \times 1$ Convolution-ReLU block to decrease the channels from $c$ to 32, followed by a $3 \times 3$ Convolution-ReLU block to augment the spatial information. Given our hypothesis that only rudimentary information about the target location is required, we audaciously further diminish the channel dimension to curtail computational overhead. $\Phi_{\text{Conv2D}}$ signifies a $3 \times 3$ Convolution-ReLU block, whereas $\Phi_{\text{FusionConv}}$ denotes the final fusion convolution layers, consisting of two $3 \times 3$ convolution layers and a $1 \times 1$ convolution layer to restore the channel dimension.

\vspace{6pt}
\noindent \textbf{Feature Splitter:} 
To procure multi-scale features, a straightforward feature splitter ($\Phi_{\text{f-split}}$) is applied to $F_{\text{agg}}$. The $\Phi_{\text{f-split}}$ employs transposed convolutional layers to generate up-sampled features, and max pooling to yield down-sampled features. By leveraging the up/down-sampling layers, we ultimately derive five feature maps of varying scales, \textit{i.e.}, $\{F_{\text{ms}}^j \in \mathbb{R}^{\frac{H}{2^{j+1}} \times \frac{W}{2^{j+1}} \times c} \}$, where $j \in \{1,2,3,4,5\}$.

\vspace{8pt}
\subsubsection{Anchor-based Prompter}

Upon the acquisition of the semantic features, it becomes feasible to utilize the prompter to generate prompt embeddings for the SAM mask decoder. This section will primarily concentrate on the exploration of the anchor-based prompter.

\begin{figure}[t]
\centering
\resizebox{\linewidth}{!}{
\includegraphics[width=\linewidth]{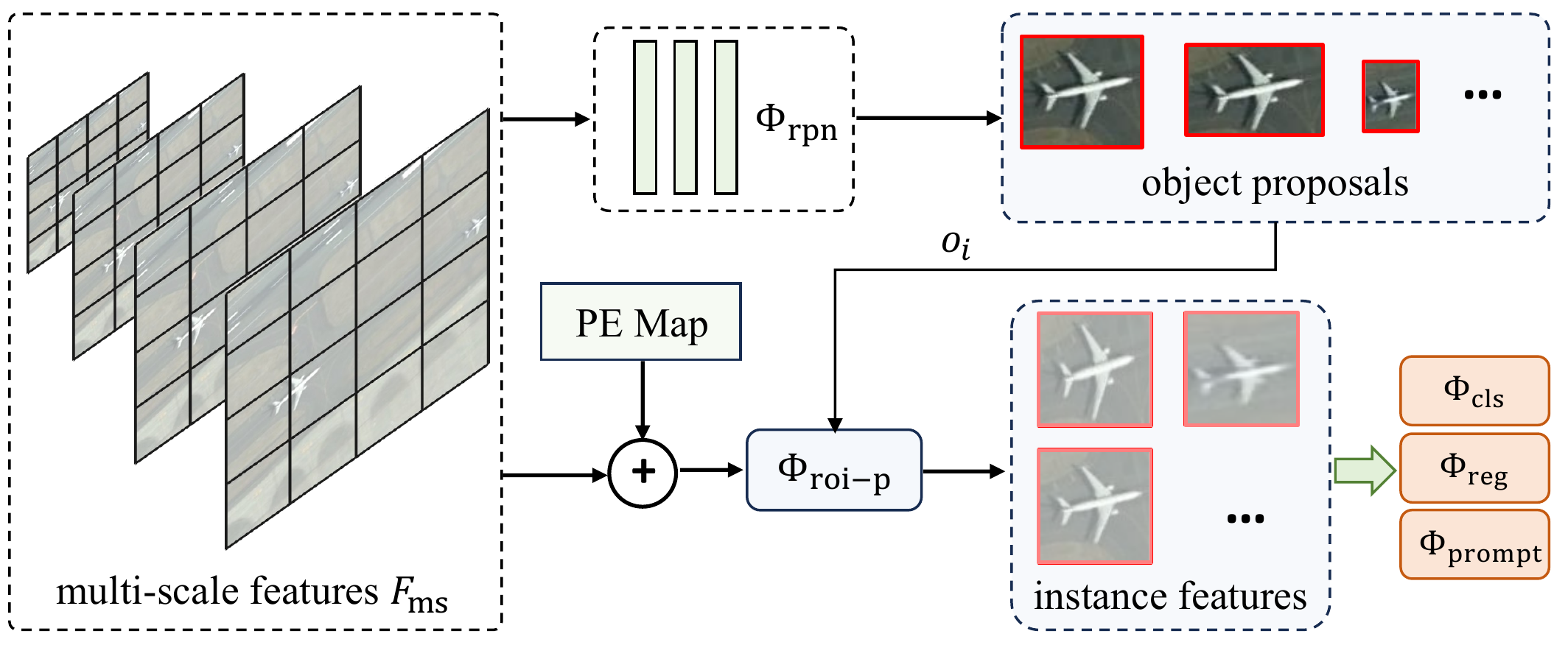}
}
\caption{
A diagram of the anchor-based prompter.}
\label{fig:anchor}
\end{figure}

\vspace{6pt}
\noindent \textbf{Architecture}:
We begin with generating candidate object boxes utilizing the anchor-based Region Proposal Network (RPN). Subsequently, we extract the unique visual feature representation of each object from the positionally encoded feature map via RoI Pooling \cite{he2017mask} with the proposal. This visual feature is then used to derive three perception heads: the semantic head, the localization head, and the prompt head. The role of the semantic head is to identify a specific object category, whereas the localization head is responsible for establishing the matching criterion between the generated prompt representation and the target instance mask, \textit{i.e.}, greedy matching based on localization (Intersection over Union, IoU). The prompt head, on the other hand, generates the necessary prompt embedding for the SAM mask decoder. A comprehensive illustration of the entire process is provided in Fig. \ref{fig:anchor} and can be mathematically represented by the following equation:
\begin{align}
\begin{split}
    \{o_i\} &= \Phi_{\text{rpn}}(F_{\text{ms}}) \\
    v_i &= \Phi_{\text{roi-p}}(F_{\text{ms}}+ \text{PE}, o_i) \\
    c_i &= \Phi_{\text{cls}}(v_i) \\
    b_i &= \Phi_{\text{reg}}(v_i) \\
    e_i &= \Phi_{\text{prompt}}(v_i) \\
    F_{\text{sparse}}^i &= e_i + \sin{e_i} \\
    \label{eq:RSPrompter_anchor}
\end{split} 
\end{align}
where $F_{\text{ms}}$ denotes the multi-scale features derived from the enhancer, while $\Phi_{\text{rpn}}$ signifies a lightweight RPN with object proposals ($\{o_i\}$). Given that the $\Phi_{\text{roi-p}}$ operations have the potential to lose positional information relative to the entire image in prompt generation, we integrate positional encoding ($\text{PE}$) into $F_{\text{ms}}$. $\Phi_{\text{cls}}$, $\Phi_{\text{reg}}$, and $\Phi_{\text{prompt}}$ correspond to the semantic head, the localization head, and the prompt head, respectively. To ensure alignment between the generated prompt embeddings and the embeddings from SAM's prompt encoder, we utilize sine function to directly generate high-frequency information, as opposed to predicting it through the network. This approach is adopted due to the inherent difficulty neural networks face in predicting high-frequency information. The efficacy of this design has been substantiated through experiments.

\vspace{6pt}
\noindent \textbf{Loss Function}:
The primary framework of the anchor-based prompter is fundamentally aligned with the structure of Faster R-CNN \cite{ren2015faster}. The comprehensive loss encompasses several components, namely, objectness loss and localization loss within the RPN, classification loss attributed to the semantic head, regression loss associated with the localization head, and segmentation loss for the SAM decoder. Therefore, the cumulative loss can be mathematically represented as follows:
\begin{align}
\mathcal{L}_{\text{anchor}} = \frac{1}{M} \sum_{i}^{M}\mathcal{L}_{\text{rpn}}^i + \frac{1}{N} \sum_{j}^{N}(\mathcal{L}_{\text{cls}}^j + \mathbbm{1}^j (\mathcal{L}_{\text{reg}}^j+ \mathcal{L}_{\text{seg}}^j))
\label{eq:loss_anchor}
\end{align}
where $\mathcal{L}_{\text{rpn}}$ represents the region proposal loss. $\mathcal{L}_{\text{cls}}$ denotes the Cross-Entropy (CE) loss, calculated between the predicted category and the target. $\mathcal{L}_{\text{reg}}$ is the SmoothL1 loss, computed based on the predicted coordinate offsets and the target offsets between the ground truth and the prior box. Additionally, $\mathcal{L}_{\text{seg}}$ signifies the binary CE loss between the SAM-decoded mask and the ground-truth instance mask, wherein the IoU of the boxes determines the supervised matching criteria. The indicator function $\mathbbm{1}$ is utilized to validate a positive match.

\vspace{8pt}
\subsubsection{Query-based Prompter}
The procedure for the anchor-based prompter is relatively complex, necessitating the use of box information for mask matching and supervised training. To simplify this process, we propose a novel query-based prompter that leverages optimal transport as its foundation.

\begin{figure}[t]
\centering
\resizebox{\linewidth}{!}{
\includegraphics[width=\linewidth]{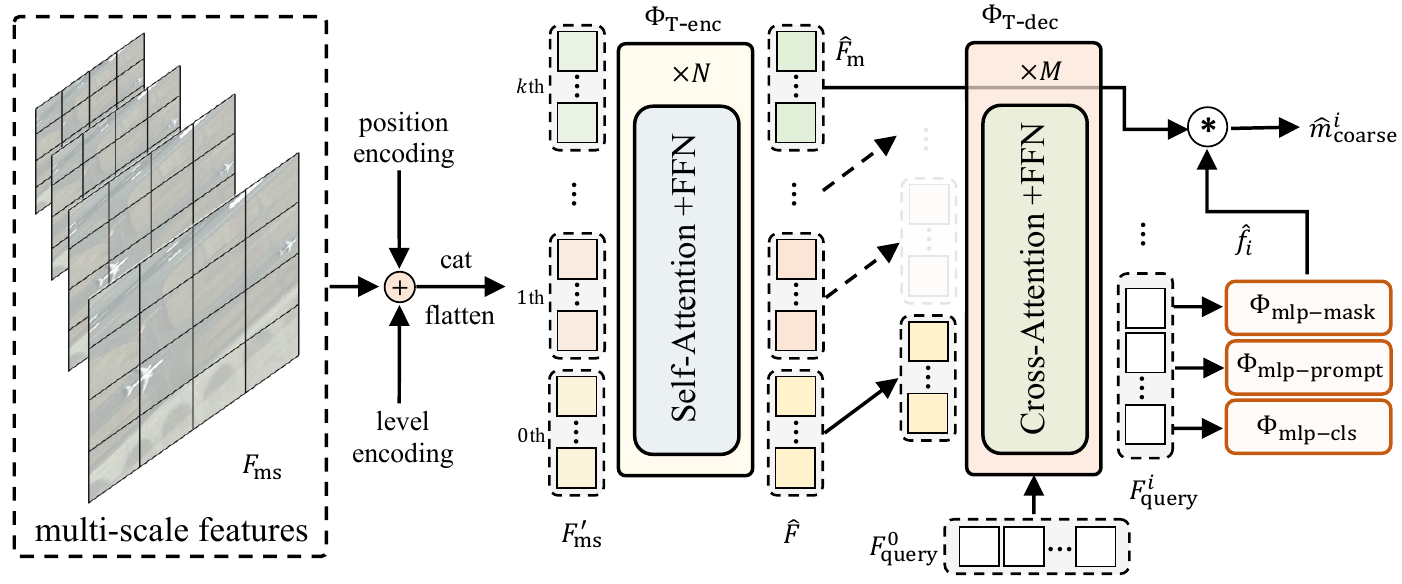}
}
\caption{
An illustration of the proposed query-based prompter.}
\label{fig:query}
\end{figure}

\vspace{6pt}
\noindent \textbf{Architecture}:
The query-based prompter is primarily composed of a Transformer encoder and decoder as shown in Fig. \ref{fig:query}. The encoder is utilized to extract high-level gathered semantic features as follows:
\begin{align}
\begin{split}
    F_{\text{ms}}^{\prime} &= \text{Cat}(\{F_{\text{ms}}^i + \text{PE}_i + \text{LE}_i\}) \\
    \hat{F_m}, \{\hat{F}_i\} &= \Phi_{\text{split}} \circ \Phi_{\text{T-enc}} (F_{\text{ms}}^{\prime}) \\
    \label{eq:RSPrompter_query_encoder}
\end{split} 
\end{align}
where $\text{PE}_i$ and $\text{LE}_i$ denote the positional encoding and level encoding of the $i$-th level, respectively. $\text{Cat}(\cdot)$ signifies tensor concatenation along the channel dimension. 
$\Phi_{\text{T-enc}}$ symbolizes the Transformer encoder layers, while $\Phi_{\text{split}}$ denotes the process of partitioning the self-attented aggregated features into $\{\hat{F}_i\}$, maintaining their original multi-scale spatial dimensions.
$\hat{F_m} \in \mathbb{R}^{\frac{H}{4} \times \frac{W}{4} \times c}$ is the largest-size feature map in $\{\hat{F}_i\}$, for coarse mask generation.

The decoder is utilized to transform the preset learnable query into the prompt embedding for SAM and corresponding semantic categories via a cross-attention interaction, as follows:
\begin{align}
\begin{split}
     \\ 
    F_{\text{query}}^i &= \Phi_{\text{T-dec}}(\hat{F}_{i-1}, F_{\text{query}}^{i-1}) \\
    \hat{c}_i &= \Phi_{\text{mlp-cls}}(F_{\text{query}}^i) \\
    \hat{f}_i &= \Phi_{\text{mlp-mask}}(F_{\text{query}}^i) \\
    \hat{e}_i &= \Phi_{\text{mlp-prompt}}(F_{\text{query}}^i) \\
    \hat{m}_{\text{coarse}}^i &= \Phi_{\text{e-sum}}(\hat{F_m}, \hat{f}_i) \\
    F_{\text{dense}}^i &= \Phi_{\text{p-enc}}(\hat{m}_{\text{coarse}}^i) \\
    F_{\text{sparse}}^i &= e_i + \sin{e_i} \\
    \label{eq:RSPrompter_query_decoder}
\end{split} 
\end{align}
where $F_{\text{query}}^0 \in \mathbb{R}^{N_p \times c}$ denotes zero-initialized learnable tokens. $\Phi_{\text{mlp-cls}}$ and $\Phi_{\text{mlp-mask}}$ are linear projection layers employed to derive the object class ($\hat{c}_i \in \mathbb{R}^{N_p \times \mathcal{C}}$) and mask filter ($\hat{f}_i \in \mathbb{R}^{N_p \times c}$) for the $i$-th level, respectively.
$\Phi_{\text{mlp-prompt}}$ is a two-layer MLP designed to acquire the projected prompt embeddings ($\hat{e}_i \in \mathbb{R}^{N_p \times K_p \times c}$). Here, $N_p$ signifies the number of prompt groups, \textit{i.e.}, the number of instances. $K_p$ is used to define the number of embeddings per prompt, \textit{i.e.}, the number of prompts required to represent an instance target.
The $i$-th level’s coarse segmentation mask is achieved by $\Phi_{\text{e-sum}}$ through linearly weighting $\hat{F_m}$ using $\hat{f}_i$. $\Phi_{\text{p-enc}}$ is the SAM mask prompt encoder. $F_{\text{dense}}^i$ and $F_{\text{sparse}}^i$ can be decoded by the SAM mask decoder as Eq. \ref{eq:RSPrompter} to yield the fine-grained mask prediction.
The overall equation can be computed recurrently along the level $i$ to obtain multi-semantic results. For inference, only the last layer is considered, and the tight bounding boxes of binary masks are procured through mathematical operations.

\vspace{6pt}
\noindent \textbf{Loss Function}:
In the training process for the query-based prompter, two primary steps are undertaken: (i) the matching of $N_p$ predicted masks to $K$ ground-truth instance masks (generally, $N_p > K$); (ii) the subsequent implementation of supervised training utilizing the matched labels. During the execution of optimal transport matching, we establish the following matching cost, which incorporates both the predicted category and mask:
\begin{align}
\begin{split}
    \Omega &= \arg \min_{\omega} \sum_i^N \mathcal{L}_{\text{match}}(\hat{y}_i, y_{\omega(i)}) \\
    \mathcal{L}_{\text{match}} &= \mathcal{L}_{\text{cls}} + \mathcal{L}_{\text{seg-ce}} + \mathcal{L}_{\text{seg-dice}} \\
    \label{eq:loss_query_match}
\end{split} 
\end{align}
where $\omega$ denotes the assignment relationship, while $\hat{y}_i$ and $y_i$ represent the prediction and the label, respectively. The Hungarian algorithm \cite{kuhn1955hungarian} is utilized to identify the optimal assignment between the $N_p$ predictions and $K$ targets. The matching cost takes into account the similarity between predictions and ground-truth annotations. More specifically, it includes the class classification matching cost ($\mathcal{L}_{\text{cls}}$), mask cross-entropy cost ($\mathcal{L}_{\text{seg-ce}}$), and mask dice cost ($\mathcal{L}_{\text{seg-dice}}$).

Upon pairing each predicted instance with its corresponding ground truth, we are then able to apply the supervision terms. These primarily consist of multi-class classification and binary mask classification, as detailed below:
\begin{align}
\begin{split}
    \mathcal{L}_{\text{query}} = \frac{1}{N_p} \sum_{i}^{N_p}(\mathcal{L}_{\text{cls}}^i + \mathbbm{1}^i \mathcal{L}_{\text{seg}}^i)
    \label{eq:loss_query}
\end{split} 
\end{align}
where $\mathcal{L}_{\text{cls}}$ represents the cross-entropy loss calculated between the predicted category and the target. $\mathcal{L}_{\text{seg}}$ denotes the binary cross-entropy loss between the predicted mask and the matched ground-truth instance mask, encompassing both the predicted coarse and fine-grained masks. $\mathbbm{1}$ is an indicator to confirm a positive match.

\subsection{Extensions on SAM for Instance Segmentation}

We conducted a comprehensive survey of the SAM community. In addition to the proposed RSPrompter, we also introduced three other SAM-based instance segmentation methods for comparative analysis. These methods are illustrated in Fig. \ref{fig:model_gallery} (a), (b), and (c), and we hope that they will catalyze future research. The methods include the application of an external instance segmentation head based on SAM's image encoder, the use of an object classifier predicated on SAM's ``everything" mode, and the incorporation of an additional object detector. In the ensuing sections, these methods will be referred to as SAM-seg, SAM-cls, and SAM-det, respectively. It is worth noting that SAM-det enjoys widespread adoption within the SAM community, while the other two methods were conceived and developed by us.

\vspace{6pt}
\subsubsection{SAM-seg}

In SAM-seg, we leverage the knowledge derived from SAM's image encoder, while keeping the cumbersome encoder frozen. We utilize the multi-scale feature enhancer, as detailed in Sec. \ref{sec:ms-enhancer}, to obtain multi-scale features that are subsequently used for the instance segmentation heads, \textit{i.e.}, the mask decoder in Mask R-CNN \cite{he2017mask} and Mask2Former \cite{cheng2022masked}. The procedure can be described as follows:
\begin{align}
\begin{split}
    \{F_i\} &= \Phi_{\text{i-enc}}(\mathcal{I}) \\
\{F_{\text{ms}}^j\} &= \Phi_{\text{ms-enhancer}}(\{F_i\}) \\ 
    \mathcal{M} &= \Phi_{\text{ext-dec}}(\{F_{\text{ms}}^j\})\\ \label{eq:sam_seg}
\end{split} 
\end{align}
where $\{F_i\} \in \mathbb{R}^{k \times h \times w \times c}, i \in \{1, \cdots, k\}$ is the multi-layer semantic feature maps derived from the ViT backbone. $\Phi_{\text{ms-enhancer}}$ refers to the multi-scale feature enhancer, as detailed in Sec. \ref{sec:ms-enhancer}. $\Phi_{\text{ext-dec}}$ denotes the externally inherited instance segmentation head, which could be, Mask R-CNN \cite{he2017mask} or Mask2Former \cite{cheng2022masked}.

\vspace{6pt}
\subsubsection{SAM-cls}

In SAM-cls, we initially employ the ``everything" mode of SAM to segment all potential instance targets within the image. This is internally accomplished by uniformly distributing points across the image and considering each point as a prompt input for an instance. Upon acquiring the masks for all instances, we can assign label to each mask utilizing a classifier. The procedure can be illustrated as follows:
\begin{align}
\begin{split}
    F_{\text{img}} &= \Phi_{\text{i-enc}}(\mathcal{I}) \\
    F^{(i,j)}_{\text{sparse}} &= \Phi_{\text{p-enc}}((x_i, y_j)) \\
    F_{\text{out}}^{(i,j)} &= \text{Cat}(T_{\text{mc-filter}}, T_{\text{IoU}}, F_{\text{sparse}}^{(i,j)}) \\
    m_{(i,j)} &= \Phi_{\text{m-dec}}(F_{\text{img}}, F_{\text{out}}^{(i,j)}) \\
    c_{(i,j)} &= \Phi_{\text{ext-cls}}(\mathcal{I}, m_{(i,j)}) \\ \label{eq:sam_cls}
\end{split} 
\end{align}
where $(x_i, y_j)$ denotes the point prompt. For every image, we employ $32 \times 32$ points to generate category-agnostic instance masks. $\Phi_{\text{ext-cls}}$ denotes the external mask classifier, and $c_{(i,j)}$ refers to the assigned category. For simplicity, we directly utilize the lightweight ResNet18 \cite{he2016deep} to label the masks. It performs classification by processing the original image patch cropped by the mask. During the cropping, we first enlarge the crop box two times and then blur non-mask areas to enhance the discriminative capability. Although the mask's classification representation could potentially be extracted from the intermediate features of the image encoder for performance improvement, we have opted not to pursue this approach to maintain simplicity in our paper. Alternatively, a pre-trained CLIP model can be leveraged, enabling SAM-cls to function in a zero-shot regime without additional training.

\vspace{6pt}
\subsubsection{SAM-det}

The SAM-det with straightforward implementation has garnered considerable attention and application within the community. Initially, an object detector is trained to pinpoint the desired targets within the image. Subsequently, the detected bounding boxes are inputted as prompts into the SAM. The comprehensive process can be drawn as follows:
\begin{align}
\begin{split}
    \{b_i, c_i\} &= \Phi_{\text{ext-det}}(\mathcal{I}) \\
    F_{\text{img}} &= \Phi_{\text{i-enc}}(\mathcal{I}) \\
    F^{i}_{\text{sparse}} &= \Phi_{\text{p-enc}}(b_i) \\
    F_{\text{out}}^{i} &= \text{Cat}(T_{\text{mc-filter}}, T_{\text{IoU}}, F_{\text{sparse}}^{i}) \\  
    m_{i} &= \Phi_{\text{m-dec}}(F_{\text{img}}, F_{\text{out}}^{i}) \\
    \\ \label{eq:sam_det}
\end{split} 
\end{align}
where $\{b_i, c_i\}$ denotes the bounding boxes and their corresponding semantic labels, as identified by the externally pre-trained object detector, $\Phi_{\text{ext-det}}$. In our study, we take the Faster R-CNN \cite{ren2015faster} as the detector.

\section{Experimental Results and Analyses}
\subsection{Experimental Dataset and Settings}
\label{sec:dataset}

We utilize three publicly available remote sensing instance segmentation datasets to validate the effectiveness of the proposed method, including the WHU building extraction dataset \cite{ji2018fully}, the NWPU VHR-10 dataset \cite{cheng2014multi, su2019object}, and the SSDD dataset \cite{zhang2021sar, su2020hq}. These datasets, which vary in terms of size, categories, and modalities, have been extensively employed in the field of remote sensing instance segmentation \cite{su2020hq, su2019object, xu2021improved, wu2020improved}.

\vspace{3pt}
\noindent \textbf{WHU} \cite{ji2018fully}: 
We employ the aerial imagery subset from the WHU building extraction dataset. This subset encompasses 8188 non-overlapping RGB images, each of $512 \times 512$ pixels. These images, captured over Christchurch, New Zealand, exhibit a spatial resolution ranging from 0.0075m to 0.3m. Following the official guidelines, we allocated 4736 images to the training set, 1036 to the validation set, and 2416 to the test set.
To procure instance annotations, we utilize the connected component analysis method from the OpenCV library, enabling the transformation of semantic segmentation into an instance segmentation format.

\vspace{3pt}
\noindent \textbf{NWPU} \cite{cheng2014multi}:
The NWPU VHR-10 dataset is a remote sensing image object detection dataset comprising ten classes: airplane, ship, storage tank, baseball diamond, tennis court, basketball court, ground track field, harbor, bridge, and vehicle. It includes 715 optical remote sensing images from Google Earth, with a spatial resolution of 0.5-2m, and 85 pan-sharpened color infrared images from the Vaihingen dataset, with a spatial resolution of 0.08m. We take 80\% of the data for training and the remaining 20\% for testing. The instance annotations provided by \cite{su2019object} are utilized for both training and evaluation.

\vspace{3pt}
\noindent \textbf{SSDD} \cite{zhang2021sar}:
The SAR Ship Detection Dataset (SSDD) comprises 1160 SAR images with a resolution range spanning from 1 to 15 meters and includes 2540 ship instances. We randomly assign 80\% of the images for training and the remaining 20\% for testing. The instance masks are annotated as per the guidelines provided by \cite{su2020hq}.

\subsection{Evaluation Protocol and Metrics}

To evaluate the performance of the proposed method, we employ the widely recognized COCO \cite{lin2014microsoft} mean average precision (mAP) metric. This metric is frequently utilized to objectively assess the effectiveness of object detection and instance segmentation methods \cite{ren2015faster, chen2023ovarnet, chen2022degraded}. A prediction is deemed a true positive when the predicted box or mask of an instance exhibits an intersection over union (IoU) with its corresponding ground truth exceeding a threshold $T$, and when its predicted category aligns. In this study, we employ $\text{AP}_\text{box}$, $\text{AP}_\text{box}^{50}$, $\text{AP}_\text{box}^{75}$, $\text{AP}_\text{mask}$, $\text{AP}_\text{mask}^{50}$, and $\text{AP}_\text{mask}^{75}$ for evaluation. $\text{AP}$ refers to metrics averaged across all 10 IoU thresholds ($0.50: 0.05: 0.95$) and all categories. A larger AP value denotes more accurate predicted instance masks and, consequently, superior instance segmentation performance. $\text{AP}^{50}$ represents the calculation under the IoU threshold of 0.50, while $\text{AP}^{75}$ embodies a stricter metric corresponding to the calculation under the IoU threshold of 0.75. Consequently, $\text{AP}^{75}$ outperforms $\text{AP}^{50}$ in the evaluation of mask accuracy, with a higher $\text{AP}^{75}$ value indicating more accurate instance masks.

\subsection{Implementation Details} 

The proposed method focuses on learning to prompt remote sensing image instance segmentation utilizing the SAM foundation model. In our experimental procedures, we employ the ViT-Huge backbone of SAM, unless specified otherwise.

\vspace{6pt}
\subsubsection{Architecture Details}

The proposed method focuses on the SAM framework, which generates multiple-choice segmentation masks for a single prompt. However, our method anticipates only a single instance mask for each learned prompt. As a result, we select the first mask as the final output. For each group of prompts, we set the number of prompt embeddings to 5, \textit{i.e.}, $K_p = 5$. In the feature aggregator, we opt to take features from every 2 layers, as opposed to every layer, to enhance efficiency.
For the anchor-based prompter, the RPN network is derived from Faster R-CNN \cite{ren2015faster}, and other hyper-parameters in the training remain consistent. For the query-based prompter, we only take the last 3 small-size feature maps from the feature splitter as the inputs to improve efficiency. We employ a 3-layer transformer encoder and a 6-layer transformer decoder, implementing multi-scale training from the outputs of the decoder at 3 different levels. We only take the last-level SAM decoded mask as the final prediction.
The number of learnable tokens could be easily determined based on the distribution of object instances in each image for efficiency, \textit{i.e.}, $N_p = 100, 70, 30$ for the WHU, NWPU, and SSDD datasets, respectively. It is worth noting that a larger number also yields satisfactory results.

\begin{table*}[!htbp] 
\centering
\caption{
Comparative analysis of the proposed methods and state-of-the-art methods on the WHU dataset, demonstrating AP(\%) for boxes and masks at various IoU thresholds.
}
\label{tab:whu_sota}
\resizebox{0.75\linewidth}{!}{
\begin{tabular}{c| *{3}{c} | *{3}{c}}
\toprule
Method 
& $\text{AP}_{\text{box}}$ & $\text{AP}_{\text{box}}^{50}$ & $\text{AP}_{\text{box}}^{75}$ 
& $\text{AP}_{\text{mask}}$ & $\text{AP}_{\text{mask}}^{50}$ & $\text{AP}_{\text{mask}}^{75}$
\\
\midrule
Mask R-CNN \cite{he2017mask} & 
66.4 & 86.3 & 76.3  & 65.6 & 87.1 & 76.7
\\
MS R-CNN \cite{huang2019mask} & 
67.7 & 87.2 & 77.1 & 66.9 & 87.5 & 77.5
\\
HTC \cite{chen2019hybrid} &
68.4 & 87.8 & 77.8 & 67.7 & 88.1 & 78.3
\\
SOLOv2 \cite{wang2020solov2} & 
- & - & -& 65.2 & 86.7 & 75.2
\\
SCNet \cite{vu2021scnet} & 
68.1 & 87.6 & 77.7 & 66.5 & 87.9 & 81.2
\\
CondInst \cite{tian2020conditional} & 
66.7 & 87.5 & 76.7 & 66.6 & 87.8 & 78.7
\\
BoxInst \cite{tian2021boxinst} & 
66.7 & 86.4 & 75.7 & 55.0 & 86.5 & 63.2
\\
Mask2Former \cite{cheng2022masked} & 
69.3 & 87.2 & 78.0  & 69.2 & 88.5 & 79.3
\\
CATNet \cite{liu2021catnet} & 
66.7 & 86.3 & 76.4 & 66.1 & 86.6 & 76.8
\\
HQ-ISNet \cite{su2020hq} & 
66.1 & 86.0 & 75.7 & 66.5 & 86.4 & 78.9
\\
\midrule
SAM-seg (Mask R-CNN) & 
70.3 & 89.8 & 81.9 & 70.1 & 89.9 & 81.0
\\
SAM-seg (Mask2Former) & 
70.7 & 88.4 & 79.1  & 71.1 & 89.5 & 81.1
\\
SAM-cls & 
46.8 & 65.7 & 53.5 & 49.3 & 71.2 & 57.6
\\
SAM-det & 
69.1 & 90.1 & 79.2 & 61.8 & 89.1 & 71.0
\\
\midrule
RSPrompter-anchor & 
71.9 & 90.9 & \textbf{82.8} & 70.4 & 90.0 & 80.5
\\
RSPrompter-query & 
\textbf{72.5} & \textbf{91.0} & 81.7 & \textbf{72.5} & \textbf{92.0} & \textbf{82.9}
\\
\bottomrule
\end{tabular}
}
\end{table*}

\begin{table*}[!htbp] 
\centering
\caption{
Comparative analysis of the proposed methods and state-of-the-art methods on the NWPU dataset, demonstrating AP(\%) for boxes and masks at various IoU thresholds.
}
\label{tab:nwpu_sota}
\resizebox{0.75\linewidth}{!}{
\begin{tabular}{c| *{3}{c} | *{3}{c}}
\toprule
Method 
& $\text{AP}_{\text{box}}$ & $\text{AP}_{\text{box}}^{50}$ & $\text{AP}_{\text{box}}^{75}$ 
& $\text{AP}_{\text{mask}}$ & $\text{AP}_{\text{mask}}^{50}$ & $\text{AP}_{\text{mask}}^{75}$
\\
\midrule
Mask R-CNN \cite{he2017mask} & 
62.3 & 88.3 & 75.2 & 59.7 & 89.2 & 65.6
\\
MS R-CNN \cite{huang2019mask} & 
62.3 & 88.6 & 73.1 & 60.7 & 88.7 & 67.7
\\
HTC \cite{chen2019hybrid} & 
63.9 & 88.9 & 75.4 & 60.9 & 88.6 & 64.4
\\
SOLO v2 \cite{wang2020solov2} & 
- & - & - &  50.9 & 77.5 & 54.1
\\
SCNet \cite{vu2021scnet} & 
60.0 & 87.5 & 69.1 & 58.1 & 87.4 & 62.0
\\
CondInst \cite{tian2020conditional} & 
62.3 & 87.8 & 73.3 & 59.0 & 88.5 & 62.8
\\
BoxInst \cite{tian2021boxinst} & 
64.8 & 89.3 & 73.0 & 47.6 & 77.2 & 51.3
\\
Mask2Former \cite{cheng2022masked} & 
57.4 & 75.5 & 63.7 & 58.8 & 83.1 & 63.5
\\
CATNet \cite{liu2021catnet} & 
63.2 & 89.0 & 73.8 & 60.4 & 89.6 & 65.5
\\
HQ-ISNet \cite{su2020hq} & 
63.5 & 89.9 & 75.0 & 60.4 & 89.6 & 64.1
\\
\midrule
SAM-seg (Mask R-CNN) & 
68.8 & 92.2 & 80.1 & 65.2 & 92.0 & 71.6
\\
SAM-seg (Mask2Former) & 
63.1 & 86.3 & 70.6 & 64.3 & 89.6 & 70.1
\\
SAM-cls & 40.2 & 57.1 & 44.5 & 44.0 & 66.0 & 47.6
\\
SAM-det & 64.2 & 89.6 & 74.6 & 51.5 & 74.8 & 54.0
\\
\midrule

RSPrompter-anchor & 
\textbf{70.3} & \textbf{93.6} & \textbf{81.0} & 66.1 & \textbf{92.7} & 70.6
\\
RSPrompter-query & 
68.4 & 90.3 & 74.0  & \textbf{67.5} & 91.7 & \textbf{74.8}
\\
\bottomrule
\end{tabular}
}
\end{table*}

\begin{table*}[!htbp] 
\centering
\caption{
Comparative analysis of the proposed methods and state-of-the-art methods on the SSDD dataset, illustrating AP(\%) for boxes and masks at various IoU thresholds.
}
\label{tab:ssdd_sota}
\resizebox{0.75\linewidth}{!}{
\begin{tabular}{c| *{3}{c} | *{3}{c}}
\toprule
Method 
& $\text{AP}_{\text{box}}$ & $\text{AP}_{\text{box}}^{50}$ & $\text{AP}_{\text{box}}^{75}$ 
& $\text{AP}_{\text{mask}}$ & $\text{AP}_{\text{mask}}^{50}$ & $\text{AP}_{\text{mask}}^{75}$
\\
\midrule
Mask R-CNN \cite{he2017mask} & 
67.7 & 95.6 & 84.9 & 64.3 & 92.6 & 80.9
\\
MS R-CNN \cite{huang2019mask} & 
67.8 & 95.3 & 85.9 & 64.9 & 93.3 & 80.4
\\
HTC \cite{chen2019hybrid} & 
69.3 & 97.1 & 85.7 & 64.1 & 94.4 & 80.6
\\
SOLO v2 \cite{wang2020solov2} &
- & - & - & 58.5 & 86.2 & 74.0
\\
SCNet \cite{vu2021scnet} & 
66.9 & 92.5 &82.5 & 64.9 & 92.6 &80.1
\\
CondInst \cite{tian2020conditional} & 
68.1 & 92.4 & 85.5 & 62.5 & 93.4 & 81.2
\\
BoxInst \cite{tian2021boxinst} &
62.8 & 96.2 & 74.7 & 45.2 & 92.3 & 35.3
\\
Mask2Former \cite{cheng2022masked} & 
62.7 & 90.7 & 75.6 & 64.4 & 93.0 & 82.4
\\
CATNet \cite{liu2021catnet} & 
67.5 & 96.8 & 80.4 & 63.9 & 93.7 & 80.1
\\
HQ-ISNet \cite{su2020hq} & 
66.6 & 95.9 & 80.2 & 63.4 & 95.1 & 78.1
\\
\midrule
SAM-seg (Mask R-CNN) & 
68.7 & 97.2 & 84.3 & 66.1 & 94.5 & 83.7
\\
SAM-seg (Mask2Former) & 
63.0 &  94.9 & 75.6 & 66.5 & 95.0 & 83.6
\\
SAM-cls & 
43.2 & 70.8 & 48.8 & 47.5 & 78.1 & 57.7
\\
SAM-det & 
70.0 & 95.8 & 85.3 & 46.0 & 93.8 & 37.0
\\
\midrule
RSPrompter-anchor & 
\textbf{70.4} & \textbf{97.7} & \textbf{86.2} & 66.8 & 94.7 & 84.0 \\
RSPrompter-query & 
66.0 &  95.6 & 78.7 & \textbf{67.3} & \textbf{95.6} & \textbf{84.3}
\\
\bottomrule
\end{tabular}
}
\end{table*}

\begin{figure*}[!htpb]
\centering
\resizebox{0.99\linewidth}{!}{
\includegraphics[width=\linewidth]{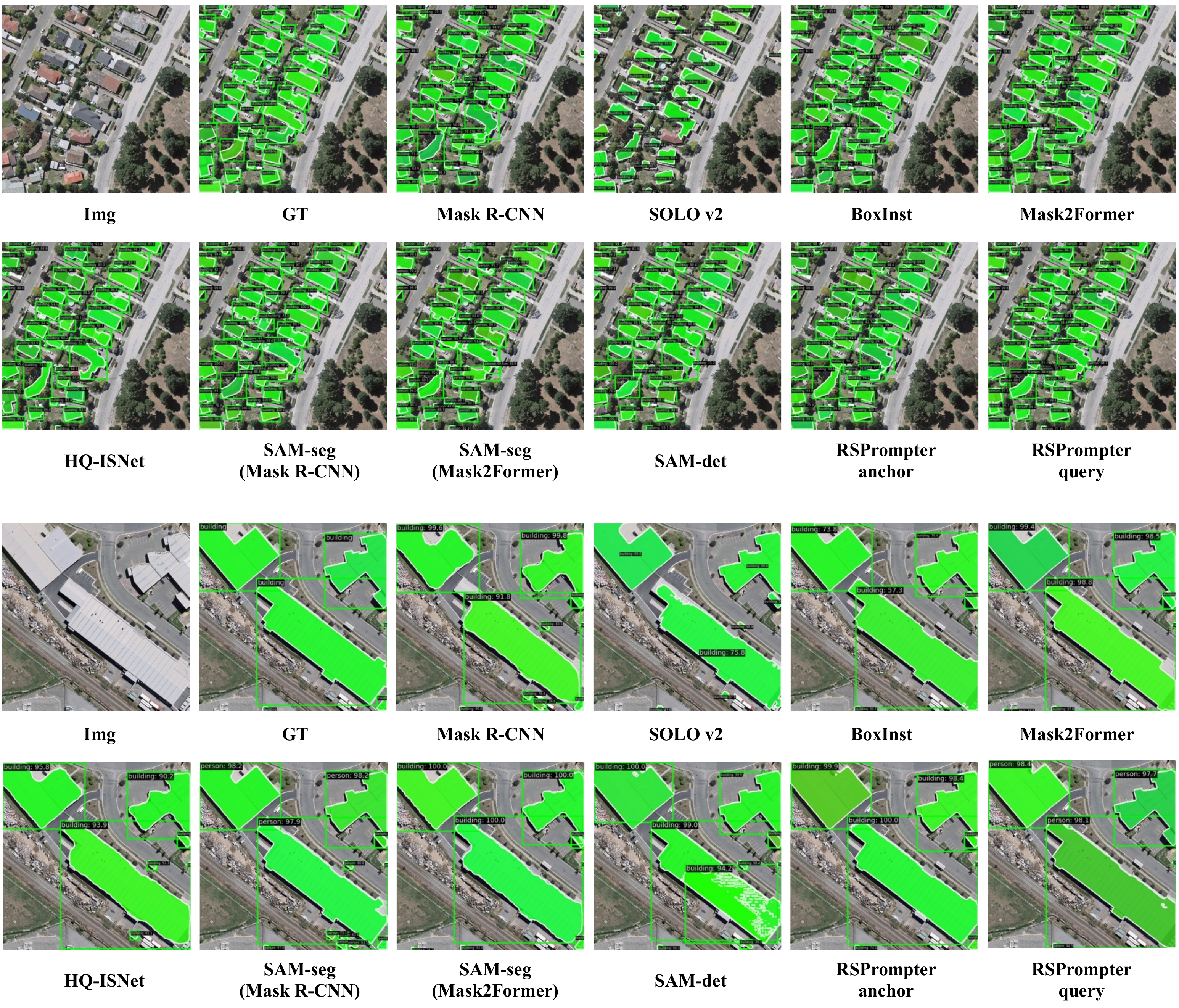}
}
\caption{
Comparative visualization of segmentation results for image samples from the WHU dataset.
}
\label{fig:vis_whu}
\end{figure*}

\begin{figure*}[!htpb]
\centering
\resizebox{0.99\linewidth}{!}{
\includegraphics[width=\linewidth]{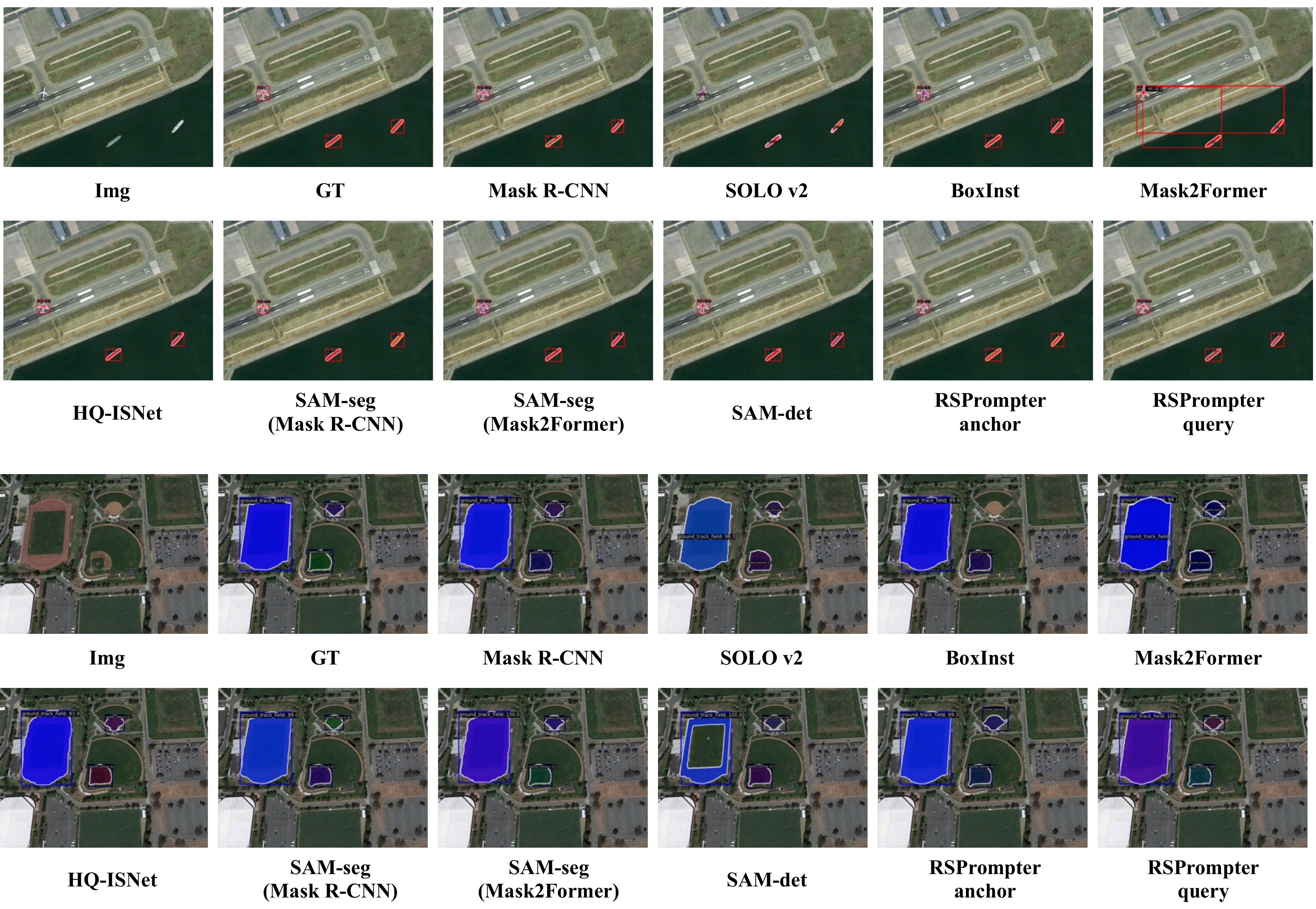}
}
\caption{
Comparative visualization of segmentation results for image samples from the NWPU dataset.
}
\label{fig:vis_nwpu}
\end{figure*}

\begin{figure*}[!htpb]
\centering
\resizebox{0.99\linewidth}{!}{
\includegraphics[width=\linewidth]{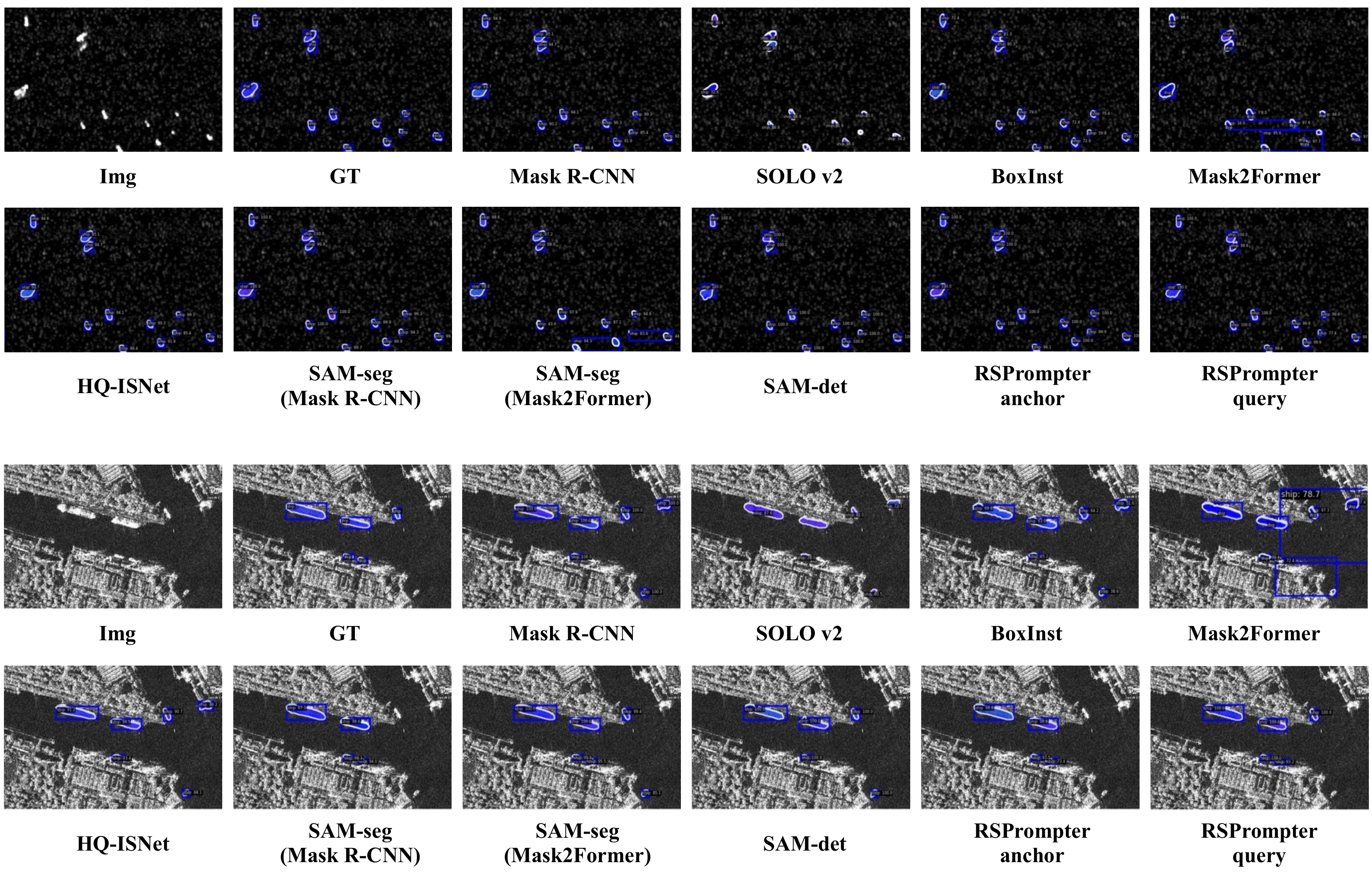}
}
\caption{
Comparative visualization of segmentation results for image samples from the SSDD dataset.
}
\label{fig:vis_ssdd}
\end{figure*}

\vspace{6pt}
\subsubsection{Training Details}

In the training phase, we adhere to an image size of $1024 \times 1024$, consistent with the original input of the SAM model. To augment the training samples, we employ techniques such as horizontal flipping and large-scale jittering. The image encoder remains frozen throughout the training phase. During the testing, we predicted up to 100 instance masks per image for evaluation. All experiments are conducted using NVIDIA A100 Tensor Core GPUs.
For the optimization process, we utilize the AdamW optimizer with an initial learning rate of $1e-4$ to train our model. We establish a mini-batch size of 16. The total number of training epochs is set at 200/300 for the WHU dataset and 300/500 for both the NWPU and SSDD datasets (RSPrompter-anchor/RSPrompter-query). We implement a Cosine Annealing scheduler \cite{loshchilov2016sgdr} with a linear warm-up strategy to decay the learning rate.
Our proposed method is developed using the PyTorch framework, and all the additional modules are trained from scratch. It is important to note that, to enhance training efficiency, we have developed and incorporated the Automatic Mixed Precision (AMP) \cite{micikevicius2017mixed} in RSPrompter-anchor training and DeepSpeed ZeRO \cite{rajbhandari2020zero} stage 2 with FP16 in RSPrompter-query.

\subsection{Comparison with the State-of-the-Art}

We compare our proposed method with a range of state-of-the-art instance segmentation methods, encompassing multi-stage approaches such as Mask R-CNN \cite{he2017mask}, Mask Scoring (MS) R-CNN \cite{huang2019mask}, HTC \cite{chen2019hybrid}, SCNet \cite{vu2021scnet}, CATNet \cite{liu2021catnet}, and HQ-ISNet \cite{su2020hq}. We also take into account single-stage methods like SOLOv2 \cite{wang2020solov2}, CondInst \cite{tian2020conditional}, BoxInst \cite{tian2021boxinst}, and Mask2Former \cite{cheng2022masked}.
Within this spectrum, SOLOv2, CondInst, BoxInst, and Mask2Former are categorized as filter-based methodologies, while CATNet and HQ-ISNet are classified as Mask R-CNN-based remote sensing instance segmentation techniques.
To augment instance segmentation methods on SAM, we have also proposed SAM-seg (Mask R-CNN) and SAM-seg (Mask2Former), which incorporate Mask R-CNN and Mask2Former heads and training regimes, respectively.
SAM-cls is envisaged as a minimalist instance segmentation method that leverages the ``everything" mode of SAM to capture all instances within the image and employs an ImageNet-initialized ResNet18 \cite{he2016deep} to label all instance masks.
SAM-det signifies the initial training of a Faster R-CNN \cite{ren2015faster} detector to procure boxes, succeeded by the generation of corresponding instance masks by SAM using the box prompts, which have been widely embraced by the community.
RSPrompter-anchor and RSPrompter-query denote the anchor-based and query-based promoters, respectively.
All the aforementioned methods are implemented following their official publications, utilizing PyTorch as the platform.

\subsubsection{Quantitative Results on the WHU Dataset}

The comparative results of RSPrompter with other methods on the WHU dataset are delineated in Table \ref{tab:whu_sota}, with the superior performance underscored in bold. The task involves executing single-class instance segmentation of buildings in optical RGB band remote sensing imagery. RSPrompter-query exhibits the most commendable performance for both box and mask predictions, achieving $\text{AP}_\text{box}$ and $\text{AP}_\text{mask}$ values of 72.5/72.5. Notably, SAM-seg (Mask2Former) outperforms the original Mask2Former (69.3/69.2) with 70.7/71.1 on $\text{AP}_\text{box}$ and $\text{AP}_\text{mask}$, while SAM-seg (Mask R-CNN) surpasses the original Mask R-CNN (66.4/65.6) with 70.3/70.1. Moreover, both RSPrompter-anchor and RSPrompter-query further enhance the performance to 71.9/70.4 and 72.5/72.5, respectively, outshining SAM-det, which conducts detection before segmentation.

These observations suggest that the learning-to-prompt approach efficaciously adapts SAM for instance segmentation tasks in optical remote sensing imagery. Furthermore, they demonstrate that the SAM backbone, trained on an extensive dataset, can offer invaluable instance segmentation guidance even when it is completely frozen (as observed in SAM-seg).

\subsubsection{Quantitative Results on the NWPU Dataset}

To further validate the effectiveness of RSPrompter, comparative experiments were conducted utilizing the NWPU dataset. Despite its smaller size relative to the WHU dataset, the NWPU dataset encompasses a more diverse range of instance categories, comprising 10 classes of remote sensing objects. The comprehensive results of various methodologies applied to this dataset are delineated in Tab. \ref{tab:nwpu_sota}. 
The results indicate that RSPrompter-anchor surpasses other methods in terms of box prediction, achieving a score of 70.3 $\text{AP}_\text{box}$. Meanwhile, RSPrompter-query excels in mask prediction, with a score of 67.5 $\text{AP}_\text{mask}$. When compared with Mask R-CNN-based methods, single-stage methods exhibit a marginal decline in performance on this dataset, particularly the Transformer-based Mask2Former. This performance decrement could potentially be ascribed to the relatively diminutive size of the dataset, which may present difficulties for single-stage methods in attaining adequate generalization across the entire data domain. This is especially true for Transformer-based methods, which require a substantial volume of training data.
Nonetheless, it is worth noting that the performance of SAM-based SAM-seg (Mask2Former) and RSPrompter-query remains commendable. The performance metrics improved from 57.4/58.8 for Mask2Former to 63.1/64.3 for SAM-seg (Mask2Former), and further to 68.4/67.5 for RSPrompter-query. 

These observations suggest that SAM, when trained on a substantial volume of data, can demonstrate significant generalization capabilities even on a smaller dataset. Despite variations in the image domain, the performance of SAM can be enhanced through the learning-to-prompt approach.

\subsubsection{Quantitative Results on the SSDD Dataset}

The SSDD dataset was utilized to conduct a comprehensive evaluation of RSPrompter's proficiency in executing remote sensing image instance segmentation. The SSDD dataset, a single-category SAR ship instance segmentation dataset, presents a distinctly different modality in comparison to the previously discussed datasets. It also exhibits substantial variations in training data from SAM. The AP values obtained for various methods on this dataset are presented in Tab. \ref{tab:ssdd_sota}. Upon analysis of the results, it is evident that the SAM-seg (Mask R-CNN) (68.7/66.1) and SAM-seg (Mask2Former) (63.0/66.5), which are predicated on the SAM backbone, yield marginal improvements over the original Mask R-CNN (67.7/64.3) and Mask2Former (62.7/64.4). This indicates a discrepancy between the image domain used for SAM training and the SAR data domain. By liberating the constrained space, both the RSPrompter-anchor and RSPrompter-query further improved the performance, thereby corroborating the effectiveness of the RSPrompter as well.

\subsubsection{Qualitative Visual Comparisons}

To facilitate a more comprehensive visual comparison with other methodologies, we conducted a qualitative analysis of the segmentation results derived from SAM-based methods and other state-of-the-art instance segmentation approaches. Fig. \ref{fig:vis_whu}, \ref{fig:vis_nwpu}, and \ref{fig:vis_ssdd} illustrate sample segmentation instances from the WHU dataset, NWPU dataset, and SSDD dataset, respectively. The proposed RSPrompter yields significant visual enhancements in instance segmentation. When compared with alternative methods, RSPrompter produces superior results, characterized by sharper edges, more defined contours, improved completeness, and a closer approximation to the ground-truth references.

\subsection{Ablation Study}

In this section, we undertake a series of experiments on the NWPU dataset to investigate the significance of each component and parameter setting within our proposed methodology. Unless stated otherwise, all models are trained to utilize the same configuration with the ViT-Huge image encoder. We confine our ablation experiments to RSPrompter-query, given its straightforward design, which led us to consider it as the principal method for this study.

\subsubsection{Impacts of Various Backbones in the Image Encoder}

Different image encoders not only affect the model's inference speed but also significantly impact its performance. SAM employs the MAE pre-trained ViT as its image encoder, which is available in three versions: base, large, and huge. We conducted experiments based on RSPrompter-query using backbones of varying parameter sizes, and Tab. \ref{tab:nwpu_ablation_backbone} presents the performance of different backbone versions on the NWPU dataset. As illustrated in the table, the instance segmentation performance exhibits an increasing trend as the size of the backbone network escalates, ranging from 64.4/65.7 to 68.4/67.5 in $\text{AP}_{\text{box}}$ and $\text{AP}_{\text{mask}}$ metrics. Depending on the specific application context, different model sizes can be selected to achieve the optimal balance between segmentation efficiency and effectiveness.

\subsubsection{Impacts of Varied Multi-scale Semantic Features in the Aggregator}

The input for the feature aggregator is sourced from features at disparate levels within the ViT backbone. To maintain optimal efficiency, we refrain from utilizing every layer of the ViT-H as the input for the feature aggregator. To demonstrate the impact of diverse feature layer selections on the ultimate segmentation results, we executed experiments delineated in Tab. \ref{tab:nwpu_ablation_feat}. The notation [start:end:step] signifies the index of the returned feature map, extending from the start to the end with the specified step size. The table demonstrates that the optimal performance and efficiency in prompt generation are achieved by features from every alternate layer.

\begin{table}[tpb] 
\centering
\caption{
Segmentation performance on the NWPU dataset with various encoders and their parameter counts.
}
\label{tab:nwpu_ablation_backbone}
\resizebox{\linewidth}{!}{
\begin{tabular}{c | c | *{3}{c} | *{3}{c}}
\toprule
Encoder & Params. & $\text{AP}_{\text{box}}$ & $\text{AP}_{\text{box}}^{50}$ & $\text{AP}_{\text{box}}^{75}$ 
& $\text{AP}_{\text{mask}}$ & $\text{AP}_{\text{mask}}^{50}$ & $\text{AP}_{\text{mask}}^{75}$
\\
\midrule
ViT-B & 86M & 64.4 & 85.3 & 71.2 & 65.7 & 90.1 & 71.1
\\
ViT-L & 307M & 66.8 & 88.8 & 72.9 & 66.0 & 91.1 & 70.5
\\
ViT-H & 632M & \textbf{68.4} & \textbf{90.3} & \textbf{74.0}  & \textbf{67.5} & \textbf{91.7} & \textbf{74.8}
\\
\bottomrule
\end{tabular}
}
\end{table}

\begin{table}[!tpb] 
\centering
\caption{
Segmentation performance with various hierarchical features to the aggregator. [start:end:step] denotes the index of the kept feature maps, ranging from the start to the end with a specified step interval.
}
\label{tab:nwpu_ablation_feat}
\resizebox{\linewidth}{!}{
\begin{tabular}{ c c | *{3}{c}|*{3}{c} }
\toprule
 Feat. & N Layer & $\text{AP}_{\text{box}}$ & $\text{AP}_{\text{box}}^{50}$ & $\text{AP}_{\text{box}}^{75}$ 
& $\text{AP}_{\text{mask}}$ & $\text{AP}_{\text{mask}}^{50}$ & $\text{AP}_{\text{mask}}^{75}$
\\
\midrule
$[0:32:1]$ & 32 & 67.9 & 89.7 & 73.7 & 67.4 & 91.5 & 73.9
\\
$ [0:32:2]$ & 16 & \textbf{68.4} & \textbf{90.3} & \textbf{74.0}  & \textbf{67.5} & \textbf{91.7} & \textbf{74.8}
\\
$ [0:32:4]$ & 8 & 66.3 & 88.5 & 72.4 & 66.3 & 90.6 & 72.0
\\
$[0:16:1]$ & 16 & 65.7 & 88.0 & 72.1 & 65.1 & 90.3 & 71.1
\\
$ [16:32:1]$ & 16 & 63.8 & 84.2 & 69.8 & 64.7 & 88.8 & 68.7
\\
$ [31:32:1]$ & 1 & 62.2 & 83.9 & 68.9 & 64.1 & 88.9 & 67.2
\\
\bottomrule
\end{tabular}
}
\end{table}

\subsubsection{Impacts of DownConv Dimension Reduction in the Aggregator}

In the pursuit of developing a lightweight feature aggregation network, we have strategically minimized the channel dimensions of the features extracted from the image encoder. Specifically, the original dimensions of 768, 1024, and 1280, as associated with ViT-B, ViT-L, and ViT-H respectively, have been reduced to a uniform dimension of 32. As evidenced in Tab. \ref{tab:nwpu_ablation_downconv_layer}, this substantial reduction in channel dimensions does not precipitate a significant decrease in performance. This suggests that it is sufficient to provide a rudimentary prompt to yield relatively precise mask results within SAM.

\begin{table}[!tpb] 
\centering
\caption{
Impact of feature dimension reduction in the aggregators on the box and mask prediction.
}
\label{tab:nwpu_ablation_downconv_layer}
\resizebox{\linewidth}{!}{
\begin{tabular}{c | *{3}{c}|*{3}{c} }
\toprule
Dimension & $\text{AP}_{\text{box}}$ & $\text{AP}_{\text{box}}^{50}$ & $\text{AP}_{\text{box}}^{75}$ 
& $\text{AP}_{\text{mask}}$ & $\text{AP}_{\text{mask}}^{50}$ & $\text{AP}_{\text{mask}}^{75}$
\\
\midrule
16 & 66.0 & 88.3 & 72.2 & 66.6 & 90.9 & 72.0
\\
32 & \textbf{68.4} & \textbf{90.3} & \textbf{74.0}  & \textbf{67.5} & \textbf{91.7} & \textbf{74.8}
\\
64 & 66.4 & 89.2 & 72.6 & 67.1 & 91.1 & 73.0
\\
128 & 64.4 & 85.2 & 70.1 & 65.9 & 89.9 & 71.5
\\
\bottomrule
\end{tabular}
}
\end{table}

\subsubsection{Impacts of Varied Architectural Designs in the Aggregator}

We have incorporated residual connections between layers to augment the propagation of semantic information within the feature aggregator, as depicted in Fig. \ref{fig:aggregator}. As corroborated by the data presented in the Tab. \ref{tab:nwpu_ablation_agg_design}, the lack of residual connections considerably affects the final segmentation efficacy, thereby implying that a wholly serial structure may not be suitable for the aggregation of features from disparate layers of ViT. Additionally, we have put forth a parallel feature aggregation structure that concatenates all the extracted feature layers for prompt generation. Nevertheless, this approach did not yield the desired performance.

\begin{table}[!tpb] 
\centering
\caption{
Segmentation performance with different architectures in the aggregator.
}
\label{tab:nwpu_ablation_agg_design}
\resizebox{\linewidth}{!}{
\begin{tabular}{ c | *{3}{c}|*{3}{c} }
\toprule
Architectures design & $\text{AP}_{\text{box}}$ & $\text{AP}_{\text{box}}^{50}$ & $\text{AP}_{\text{box}}^{75}$ 
& $\text{AP}_{\text{mask}}$ & $\text{AP}_{\text{mask}}^{50}$ & $\text{AP}_{\text{mask}}^{75}$
\\
\midrule
Ours & \textbf{68.4} & \textbf{90.3} & \textbf{74.0}  & \textbf{67.5} & \textbf{91.7} & \textbf{74.8}
\\
w/o res. connection & 62.5 & 83.5 & 69.3 & 63.8 & 87.3 & 67.9
\\
w/ para. concatenation & 63.7 & 85.0 & 70.1 & 64.9 & 89.0 & 69.7
\\
\bottomrule
\end{tabular}
}
\end{table}

\subsubsection{Impacts of Query Numbers and Prompt Embedding Numbers in the Prompter}

The prompter is designed to generate $N_p$ sets of prompts for each image, with each set symbolizing an instance mask. Each set comprises $K_p$ prompts, where $K_p$ can be interpreted as the number of points representing a target instance. These two parameters have the potential to influence the ultimate segmentation outcomes, thereby necessitating controlled experiments, as demonstrated in Tab. \ref{tab:nwpu_ablation_nprompt}. We evaluated $N_p$ values of 50, 70, and 100 while maintaining $K_p$ at 1, 3, 5, 7, and 9. The findings suggest that optimal performance is achieved when $N_p = 70$ and $K_p = 5$. To further comprehend the design of these parameters, we examined the distribution of instance numbers within the dataset and discovered that it mirrors the value of $N_p$. Consequently, we propose that the selection of $N_p$ should consider the number of targets per image present in the dataset. The choice of $K_p$ should neither be excessively small nor large: a small value may not adequately represent complex instances, while a large value may deviate from the distribution of the original number of SAM prompts. Similar trends were observed in the other two datasets as well.

\begin{table}[!tpb] 
\centering
\caption{
Segmentation performance with varying query numbers and prompt embedding numbers.
}
\label{tab:nwpu_ablation_nprompt}
\resizebox{\linewidth}{!}{
\begin{tabular}{ c c | *{3}{c}|*{3}{c} }
\toprule
 $N_p$ & $K_p$ & $\text{AP}_{\text{box}}$ & $\text{AP}_{\text{box}}^{50}$ & $\text{AP}_{\text{box}}^{75}$ 
& $\text{AP}_{\text{mask}}$ & $\text{AP}_{\text{mask}}^{50}$ & $\text{AP}_{\text{mask}}^{75}$
\\
\midrule
50 & 5 & 67.7 & 88.9 & 73.6 & 66.8 & 91.4 & 71.7
\\
70 & 5 & \textbf{68.4} & \textbf{90.3} & \textbf{74.0}  & \textbf{67.5} & \textbf{91.7} & \textbf{74.8}
\\
100 & 5 & 
66.2 & 87.7 & 72.4 & 66.6 & 90.1 & 73.8
\\
\midrule
70 & 1 & 
64.8 & 86.3 & 70.3 & 65.1 & 89.4 & 69.8
\\
70 & 3 & 
66.7 & 88.6 & 71.3 & 66.5 & 91.5 & 71.4
\\
70 & 5 & 
\textbf{68.4} & \textbf{90.3} & \textbf{74.0}  & \textbf{67.5} & \textbf{91.7} & \textbf{74.8}
\\
70 & 7 & 
66.6 & 88.8 & 72.2 & 66.9 & 89.9 & 73.3
\\
70 & 9 & 
65.2 & 87.7 & 72.5 & 65.8 & 88.6 & 72.5
\\
\bottomrule
\end{tabular}
}
\end{table}

\subsubsection{Impacts of Applying Sine Regularization for Prompt Embedding Generation in the Prompter}

The original SAM prompt encoder translates a coordinate-based prompt into high-frequency embeddings, which subsequently govern the decoding of masks through Fourier encoding. However, the feature generated by the prompter is inherently smooth due to the intrinsic characteristics of neural networks. To reconcile the prompt embeddings from both sources, we utilize a sine function to directly map the output of the prompter into the high-frequency domain. The efficacy of this design is substantiated in the second row of Tab. \ref{tab:nwpu_ablation_sine_mask_multi}. The empirical results reveal that in the absence of the sine transformation, the performance metrics decline from 68.4/67.5 to 63.7/64.6 in $\text{AP}_{\text{box}}$ and $\text{AP}_{\text{mask}}$ respectively.

\subsubsection{Impacts of Freezing the Mask Decoder}

The mask decoder in SAM is characterized by a lightweight design, and we have incorporated it into the training process. To investigate the potential enhancements, we conducted an ablation study to exclude it from training, with the results presented in the third row of Tab. \ref{tab:nwpu_ablation_sine_mask_multi}. The result suggests a significant decline in segmentation performance, thereby indicating that fine-tuning the SAM decoder for a downstream task is recommended.

\subsubsection{Impacts of Multi-scale Supervision Provided by the Splitter}

Considering that ViT generates features of uniform scale, we have designed a simplified feature splitter to provide multi-scale feature maps for ensuing multi-scale supervision during the decoding process, as depicted in Eq. \ref{eq:RSPrompter_query_decoder}). However, during the forward pass, we exclusively employ the output from the terminal layer of the decoder. The effectiveness of this design is underscored in the final row of Tab. \ref{tab:nwpu_ablation_sine_mask_multi}, where the performance metrics exhibited an improvement from 59.9/62.3 to 68.4/67.5.

\begin{table}[!tpb] 
\centering
\caption{
Impacts of the sine regularization in the prompter, training with SAM's mask decoder frozen, and employing a multi-scale training regime.
}
\label{tab:nwpu_ablation_sine_mask_multi}
\resizebox{\linewidth}{!}{
\begin{tabular}{ c | *{3}{c}|*{3}{c} }
\toprule
Ablation & $\text{AP}_{\text{box}}$ & $\text{AP}_{\text{box}}^{50}$ & $\text{AP}_{\text{box}}^{75}$ 
& $\text{AP}_{\text{mask}}$ & $\text{AP}_{\text{mask}}^{50}$ & $\text{AP}_{\text{mask}}^{75}$
\\
\midrule
Ours (none) & 
\textbf{68.4} & \textbf{90.3} & \textbf{74.0}  & \textbf{67.5} & \textbf{91.7} & \textbf{74.8}
\\
w/o sine & 63.7 & 85.0 & 69.8 & 64.6 & 88.1 & 70.2
\\
w/o train dec. & 56.8 & 77.5 & 62.0 & 59.5 & 83.0 & 63.8
\\
w/o ms loss & 59.9 & 80.2 & 66.6 & 62.3 & 85.8 & 67.7
\\
\bottomrule
\end{tabular}
}
\end{table}

\subsubsection{Impacts of Varying Knowledge Bases on SAM-seg}

We regard models pre-trained on extensive datasets as knowledge bases. In this context, we have executed experiments on SAM-seg (Mask2Former) to evaluate the influence of various pre-trained knowledge bases on remote sensing segmentation. Unless explicitly mentioned, the model architecture and training regimes align with those of SAM-seg (Mask2Former), implying that the image encoder is frozen. 
We have conducted experiments employing the following methods: ResNet50 pre-trained on ImageNet \cite{he2016deep}, ViT-B pre-trained with MAE \cite{he2022masked}, SatMAE \cite{cong2022satmae}, Scale-MAE \cite{reed2023scale}, SeCo \cite{manas2021seasonal}, and CACo \cite{mall2023change}. As presented in Tab. \ref{tab:nwpu_ablation_knowlwdge_base}, the experimental results suggest that the latent knowledge offered by the SAM backbone is more apt for segmentation tasks.

\begin{table}[!tpb] 
\centering
\caption{
Impacts of varying knowledge bases on instance segmentation performance. All the backbones are frozen except ImageNet pre-trained R50.
}
\label{tab:nwpu_ablation_knowlwdge_base}
\resizebox{\linewidth}{!}{
\begin{tabular}{ l c | *{3}{c}|*{3}{c} }
\toprule
Base & Backbone & $\text{AP}_{\text{box}}$ & $\text{AP}_{\text{box}}^{50}$ & $\text{AP}_{\text{box}}^{75}$ 
& $\text{AP}_{\text{mask}}$ & $\text{AP}_{\text{mask}}^{50}$ & $\text{AP}_{\text{mask}}^{75}$
\\
\midrule
ImageNet \cite{he2016deep} & R50$^*$ & 57.4 & 75.5 & 63.7 & 58.8 & 83.1 & 63.5
\\
MAE \cite{he2022masked} & ViT-B & 49.3 & 71.3 & 53.3 & 51.6 & 76.7 & 54.1
\\
SeCo \cite{manas2021seasonal} & R50 & 55.5 & 74.4 & 61.0 & 56.7 & 79.8 & 60.1
\\
SatMAE \cite{cong2022satmae} & ViT-L & 57.2 & 78.3 & 63.2 & 59.6 & 84.0 & 63.6
\\
Scale-MAE \cite{reed2023scale} & ViT-L & 59.6 & 79.3 & 66.4 & 60.1 & 83.7 & 63.7
\\
CACo \cite{mall2023change} & R50 & 57.8 & 79.1 & 64.1 & 59.0 & 84.2 & 62.7
\\
SAM (Ours) & ViT-H & \textbf{63.1} & \textbf{86.3} & \textbf{70.6} & \textbf{64.3} & \textbf{89.6} & \textbf{70.1}
\\
\bottomrule
\end{tabular}
}
\end{table}

\subsubsection{Analysis of the Low Performance on SAM-cls}

SAM-cls is entirely dependent on SAM for the instance segmentation of remote sensing images, which is subsequently followed by the training and annotation of a classifier on the predicted segmentation results. As evinced in the preceding experiments, it manifests sub-optimal performance. To negate the influence engendered by the performance of the classifier, we exhibit the Top-1 accuracy during classification. The result is 98.5\%, which underscores that the classifier is capable of accurately categorizing various classes, which further suggests that the low performance of SAM-cls can be ascribed to the pre-segmentation conducted by SAM due to the domain shift.

\subsubsection{Oracle Experimentation on SAM-det}

\begin{figure}[t]
\centering
\resizebox{\linewidth}{!}{
\includegraphics[width=\linewidth]{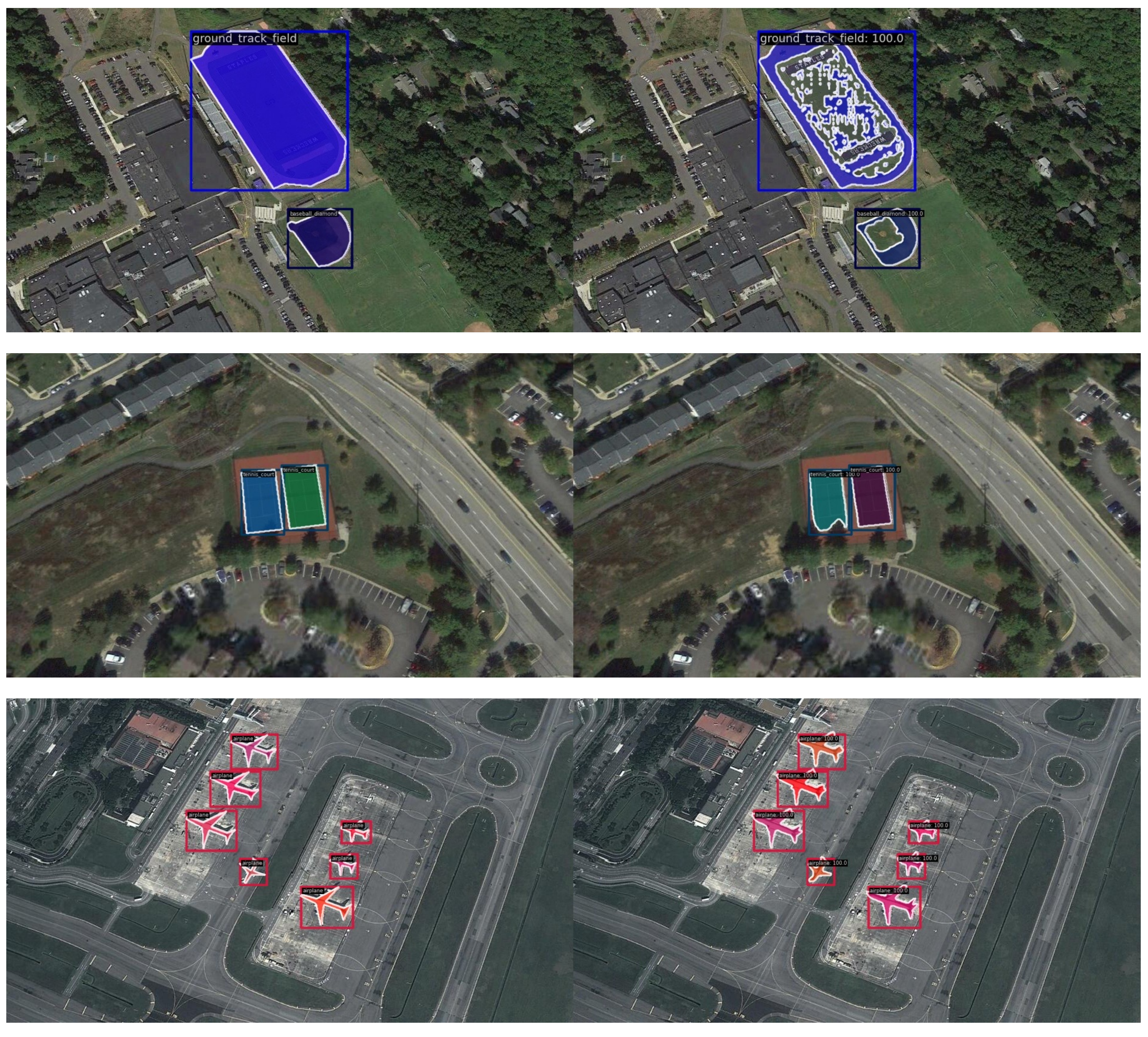}
}
\caption{
Some results from SAM-det (oracle). Left: ground-truths; Right: predictions.
}
\label{fig:samdet_vis}
\end{figure}

SAM-det utilizes a detector to procure box-based prompts, which are subsequently incorporated into SAM for mask extraction. Within this framework, we directly input annotated boxes to ascertain the maximum potential of SAM-det. It becomes apparent that despite the accuracy of the categories and the precision of the box-level annotations, SAM-det does not consistently deliver robust segmentation results, as shown in Tab. \ref{tab:nwpu_samdet}. This observation implies a substantial discrepancy when SAM is directly employed in the perception of remote sensing scenarios. Furthermore, the provision of box-level prompts does not invariably ensure the completeness of the mask. Fig. \ref{fig:samdet_vis} illustrates examples of various segmentation scenarios.

\begin{table}[!htbp] 
\centering
\caption{
A considerable performance margin, even leveraging the annotated boxes for SAM.
}
\label{tab:nwpu_samdet}
\resizebox{1\linewidth}{!}{
\begin{tabular}{c| *{3}{c} | *{3}{c}}
\toprule
Method 
& $\text{AP}_{\text{box}}$ & $\text{AP}_{\text{box}}^{50}$ & $\text{AP}_{\text{box}}^{75}$ 
& $\text{AP}_{\text{mask}}$ & $\text{AP}_{\text{mask}}^{50}$ & $\text{AP}_{\text{mask}}^{75}$
\\
\midrule
box-predicted & 64.2 & 89.6 & 74.6 & 51.5 & 74.8 & 54.0
\\
box-annotated & - & - & - & 52.6 & 84.6 & 55.8 
\\
\bottomrule
\end{tabular}
}
\end{table}

\subsection{Discussions}

In this paper, we introduce a prompt-learning methodology based on SAM, designed to enhance the processing of remote sensing images with foundation models. This method is not confined to the SAM model and can be extrapolated to a range of foundation models. Throughout the experimental phase, we have pinpointed several potential areas for enhancement.

As highlighted by SAM, while the mask decoder of SAM is notably lightweight (4.1M), it does not necessarily denote that its computational demands are similarly minimal. The input token number of the mask decoder transformer is considerable ($> 64 \times 64$), and as a prompt-based interactive segmentation head, a forward calculation is necessitated for each prompt group. Therefore, when addressing 100 instance targets within a single image, the forward calculation must be executed 100 times, which is computationally demanding. Researchers may contemplate reconfiguring this segmentation head for downstream tasks. Additionally, RSPrompter-query, predicated on optimal transport matching, converges at a slower pace than RSPrompter-anchor due to the absence of relatively explicit supervisory information. However, its network structure is more simplistic, lightweight, and exhibits superior performance compared to RSPrompter-anchor. Researchers could investigate strategies to optimize its convergence speed. Finally, the prompt learning methodology proposed herein demonstrates exceptional generalization performance on smaller datasets, significantly outperforming alternative methods. This indicates that in situations where data is insufficient to train or fine-tune an appropriate network, prompt engineering design could be contemplated for foundation models.
\section{Conclusion}

In this paper, we present RSPrompter, a novel prompt learning methodology for instance segmentation of remote sensing images, which capitalizes on the SAM foundation model. The primary objective of RSPrompter is to learn the generation of prompt inputs for SAM, thereby enabling it to autonomously procure semantic instance-level masks. This stands in contrast to the original SAM, which necessitates additional manually-crafted prompts to attain category-agnostic masks. The design philosophy underpinning RSPrompter is not confined to the SAM model and can be extrapolated to other foundational models. Based on this philosophy, we have developed two specific implementation schemes: RSPrompter-anchor, predicated on pre-set anchors, and RSPrompter-query, which is reliant on queries and optimal transport matching. Each structure harbors its unique merits. Additionally, we have explored and proposed a variety of methods and variants within the SAM community for this task and compared them with our prompt learning approach. The efficacy of each component within the RSPrompter has been validated through comprehensive ablation studies. Simultaneously, experimental results on three public remote sensing datasets substantiate that our method surpasses other state-of-the-art instance segmentation methods, as well as several additional SAM-based methods.

\ifCLASSOPTIONcaptionsoff
  \newpage
\fi

\bibliographystyle{IEEEtran}
\bibliography{IEEEabrv,myreferences}

\end{document}